\documentclass{article}

\usepackage{iclr2025_conference,times}

\usepackage{natbib}

\usepackage{amsmath,amsfonts,bm}

\def\eqref#1{equation~\ref{#1}}
\def\1{\bm{1}}

\DeclareMathAlphabet{\mathsfit}{\encodingdefault}{\sfdefault}{m}{sl}
\SetMathAlphabet{\mathsfit}{bold}{\encodingdefault}{\sfdefault}{bx}{n}

\usepackage{hyperref}
\usepackage{url}
\usepackage{graphicx}
\usepackage{booktabs}
\usepackage{enumitem}
\usepackage{tabularx}

\iclrfinaltrue
\title{Emergent properties with repeated examples}
\author{Fran\c{c}ois Charton\\
FAIR, Meta\\
\texttt{fcharton@meta.com} \\
\And
Julia Kempe\\
FAIR, Meta \& NYU CDS and Courant Institute\\
\texttt{kempe@meta.com} \\
}

\begin{document}

\maketitle

\begin{abstract}
{We study the performance of transformers as a function of the number of repetitions of training examples with algorithmically generated datasets. On three problems of mathematics: the greatest common divisor, modular multiplication, and matrix eigenvalues, we show that for a fixed number of training steps,
models trained on smaller sets of repeated examples outperform models trained on larger sets of single-use examples. We also demonstrate that {\em two-set training} - repeated use of a small random subset of examples, along normal sampling on the rest of the training set - provides for faster learning and better performance.
This highlights that the benefits of repetition can outweigh those of data diversity. 
These datasets and problems provide a controlled setting to shed light on the still poorly understood interplay between generalization and memorization in deep learning. }
%Pour l'enfant, amoureux de cartes et d'estampes, \\
%L'univers est égal à son vaste appétit. \\
%Ah! Que le monde est grand à la clarté des lampes! \\Aux yeux du souvenir que le monde est petit!
\end{abstract}

\section{Introduction}

When training neural networks, 
it has become customary to use  
the largest and most diverse datasets available, and to limit example reuse as much 
as possible. This tendency is manifest in large language models. 
GPT~\citep{Radford2018ImprovingLU} was trained for $100$ epochs (each example was seen $100$ times on average), BERT~\citep{BERT} on $40$ and GPT-2~\citep{radford2019language} on $20$. In recent models, most examples in the pre-training corpus are seen only once, a few specialized datasets are iterated $2$ or $3$ times, and fine-tuning examples are seen once or twice. Meanwhile, data budgets are on the increase:
GPT-2 was trained on less than $10$ billion tokens, GPT-3~\citep{GPT3brown2020} was pre-trained on $300$ billion, Chinchilla~\citep{chinchillahoffmann2022} and Llama~\citep{llamatouvron2023} on $1.4$ trillion, Llama2~\citep{llama2touvron2023} on $2$ trillion, and Llama3~\citep{llama3_2024} on $15.6$ trillion. 
Whereas the use of large train sets is grounded in theory~\citep{vapnik82}, the practice of not repeating training examples is less motivated. It reflects the belief that, when availability permits fresh data is superior to repeated use of a corpus~\citep{komatsuzaki2019epochneed,RAVIV2022VariabilityCogSci, hernandez2022scalinglawsinterpretabilitylearning, muennighoff2023fourepochs}. 
This belief is grounded in the idea that memorization of repeated examples 
hinders generalization \citep{zhang2017understanding}.  From a human learner point of view, this is counter-intuitive. When faced with a situation we never experienced, we \emph{recall} similar instances~\citep{ProustSwann}, and use them as anchors to navigate the unknown. If memorization benefits human learners~\citep{CogScirepetitio2015}, why should it hinder machines?

In this paper we challenge the view that the repetition of training examples is undesirable, 
and that for a given \emph{training budget} 
(TB, the \emph{total} number of training examples), one should maximize the \emph{data budget} (DB, the number of \emph{distinct} training examples). We explore the impact of repeated samples in three controlled settings using generated data: computing the greatest common divisor (GCD) of two integers~\citep{charton2023gcd}, modular multiplication of two integers, and calculating the eigenvalues of symmetric real matrices~\citep{charton2022linear}. 
These settings allow for perfect control over the distribution of repeated examples, unlike \emph{natural datasets} (e.g. text from the web) which may feature unintended duplication and redundancy.
\begin{figure}[t]
\begin{minipage}{0.5\textwidth} 
\includegraphics[width=\textwidth]{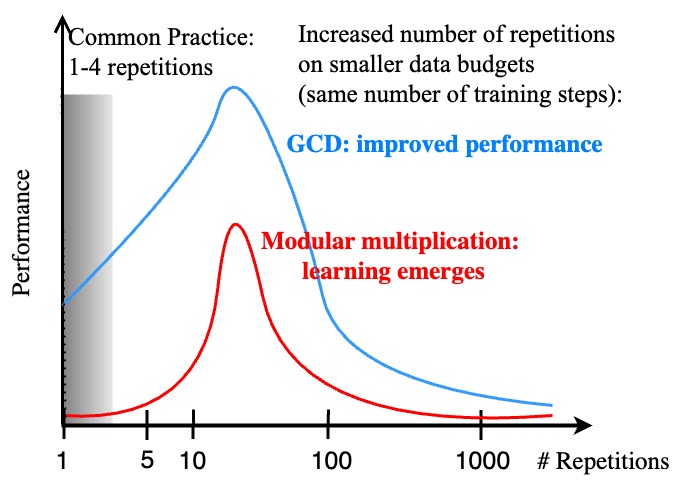}%\label{fig:teaser1}
\end{minipage}\hfill 
\begin{minipage}{0.5\textwidth}
\includegraphics[width=.9\textwidth]{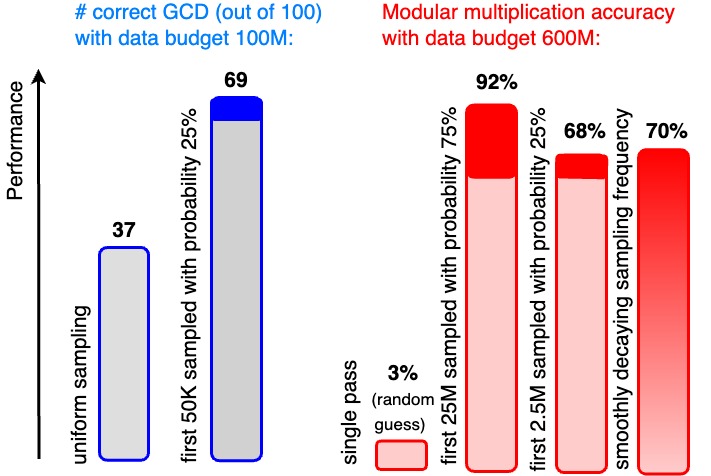}%\label{fig:teaser2}
\end{minipage}
\caption{\small{{\bf Repetition Helps (Left):} Performance as a function of repetition for a fixed training budget ($600$M). {\em GCD (blue)}. Models trained on smaller datasets, repeated $30$ times, perform much better than models trained on one to four epochs. {\em Multiplication mod 67 (red)}. Models trained for $1$ to $4$ epochs do not learn. Learning ``emerges'' when models are trained on smaller data budgets, with increased repetition.\\ 
{\bf Two-set training (Right):}  For a fixed data budget, splitting the data into two {\em random} subsets and increasing the training frequency of one greatly improves performance. {\em GCD (left)}: repeating 50k examples 3000 times for a training budget of 600M brings performance from 37 to 69 on 100M.  {\em Modular multiplication (right)}: Models trained on 600M single-use examples do not learn. With $25$M examples repeated $18$ times, and $150$M single use examples, accuracy is $92\%$, with $2.5$M examples repeated $60$ times, and $450$M single-use, accuracy is $68\%$. Smooth distributions of repetition over the training set achieve $70\%$ accuracy.}}
  \label{fig:teaser}
  \vspace{-0.5cm}
\end{figure}
Our experiments uncover two striking phenomena:
\begin{enumerate}[nosep,noitemsep, topsep=0pt, leftmargin=15pt]
    \item {\bf Repetition Helps:} For fixed training budgets (300M to 1B examples), models trained from small data budgets (25 to 50M examples) outperform models trained on large DB. This sometimes gives rise to ``emergent'' phenomena: properties \emph{only learned} by models trained on small DB.
    \item {\bf Two-Set Training% - Divide and Conquer
    :} For fixed data budgets, learning speed and performance are significantly enhanced by \emph{randomly selecting} a subset of training examples, and repeating them more often during training.  The ``two-set effect" is all the more surprising as the repeated examples are not curated, and only differ from the rest of the training data by their frequency of use. 
    \end{enumerate}

In ablation experiments, we show that the performance of two-set training cannot be improved by curating the set of repeated examples, or refreshing it as training proceeds. This sets us apart from \emph{curriculum learning}, and strengthens the observation that repetition of a few \emph{random examples} is really all we need. We also show that mixing repeated and non-repeated examples in the same mini-batches is required for two-set training to work. Finally, we propose a smooth extension of two-set training, by introducing a probability distribution on the training set.

Our work isolates an interesting phenomenon in a clean setting. The three tasks we study each exhibit idiosyncratic structure that allows to test a variety of hypotheses. For instance, the GCD dataset exhibits an inverse polynomial distribution of results, reminiscent of Zipf's law  in natural language \citep{zipf1935psycho}. This allows us to test whether amplification of the tail of the distribution can benefit learning, by incorporating it into two-set training (while an attractive hypothesis, our ablations show that this seems not to be the case). In contrast, the results of modular multiplication are almost uniformly distributed, indicating that our conclusions do not depend on the existence of a power-law. Finally, the eigenvalue problem features non-linear, approximate calculations on reals. 

In all three cases, the benefits of repetition are significant, but come in different flavors, from improving performance and accelerating learning (GCD), to allowing a new task to be learned (multiplication), or to be accessible to smaller models (eigenvalues). Alternatively, small random subsets of the data repeated at high frequency can elicit similar effects. These findings have profound implications and should lead to a paradigm shift where the training set size becomes a mere hyper-parameter, not solely governed by the availability of data and the belief that more is always better. 

\textbf{Note.} \emph{Training budget} is known as {\em compute budget} in other works ~\citep{Power2022Grokking,muennighoff2023fourepochs}. We use {\em training budget} to distinguish it from the compute cost arising from model size.

\section{Background and Related Work}

In this paper, we focus on relatively small transformer models performing mathematical tasks, placing it into a long established corpus of works that study interesting phenomena in a controlled setting, and advance our understanding of the underlying mechanisms in larger models in the wild, see e.g. \citet{Power2022Grokking,NEURIPS2022LiangInContext,charton2023gcd,dohmatob2024aTaleTails}. 

One such example is the study of {\em ``grokking"}, first observed with modular arithmetic - a phenomenon where models generalize long after achieving $100\%$ accuracy on their (small) training set \citep{Power2022Grokking,NEURIPS2022grok,liu2023omnigrok}. On the surface, grokking shares similarities with our work: a small training dataset is iterated for many epochs, the phenomenon is isolated in clean experiments on synthetic data, and it contradicts traditional wisdom regarding overfitting \citep{MohriLearningBook2018}. But there are important differences: in grokking, delayed learning occurs, we observe no such delay; grokking occurs for ``tiny'' training samples (hundreds or thousands of examples), our models use millions (even for modular multiplication); grokking is very sensitive to the optimizer used, our findings are robust across optimizers (Appendix~\ref{app:optimizers}), and, of course, no two-set approach is documented in the grokking setting. 

Another related setting is {\em ``benign overfitting"} \citep{BenignPNAS2020,Belkin_2021benign,Bartlett_Montanari_Rakhlin_2021benign}, where an {\em over-parametrized} model perfectly fits noisy data, without harming prediction accuracy. One could argue that our work presents a \emph{quantitative} manifestation of benign overfitting, inasmuch as decreasing the data budget increases model over-parametrization. However, this would not account for the decrease in performance once the data budget falls below a certain number (one could argue that overfitting is no longer benign, then), nor for the possibility of two-set training.

Prior works have studied the role of data reuse in language models.  
\citet{hernandez2022scalinglawsinterpretabilitylearning} study data repetition in models with up to $800$M parameters with training budgets of $100$M tokens to exhibit detrimental impact of repetition of a subset of the training data. In the context of scarcity of training data, \citet{muennighoff2023fourepochs} find for LLMs of contemporary size (up to $9$B)
that with constrained data for a fixed training
budget, training with up to $4$ epochs of repeated data yields negligible changes to
loss compared to having unique data; any further repetition decreases the value of additional training.
A limitation of these works is the lack of control over repetition in the training set: partial copies, of sentences, paragraphs sometimes whole documents, abound in pre-training corpora. \citet{allen-zhu2024physics} undertake a controlled study on synthetic language data in the context of knowledge {\em retrieval} and find that knowledge augmentation - repeated inclusion of reformulated variants - of a small subset of the data leads to performance improvement; an effect somewhat akin to what we observe in two-set training.

Our work is related to, but different from, {\em curriculum learning (CL)} \citep{BengioCurriculum2009,WangCurriculumSurvey2021}, where training data is presented in a meaningful order, usually from ``easy" to ``hard" samples. Two-set training differs from curriculum learning in at least two important ways: in CL, datasets are curated, our subsets are completely random; in CL, the training distribution shifts over time, while our subsets are static. Our ablations show that curating the repeated set, or changing it over time, as in CL, brings no improvement in performance (and may even have an adverse effect).

Lastly, our work touches upon the expansive area of {\em out-of-distribution (OOD)} generalization \citep{gulrajani2021:ood,tipLopePaz}, which studies generalization when train and test distributions differ. Curiously, while our two-set approach increases the frequency of some training examples, because the repeated set is chosen {\em at random}, the training set remains distributionally equivalent to the test set. Thus, our study falls outside the usual framework of OOD studies.

\section{Experimental settings and baselines}

We focus on three problems of mathematics: computing the greatest common divisor, multiplication modulo $67$, and computing the eigenvalues of real symmetric matrices. The GCD and eigenvalues were studied in prior work~\citep{charton2022linear,charton2023gcd,dohmatob2024aTaleTails,feng2024beyond}.

\textbf{Greatest common divisor.} 
The model is tasked to predict the GCD of two integers uniformly distributed between $1$ and $1$ million, encoded in base $1000$. Following \citet{charton2023gcd}, who observes that throughout training almost all pairs of integers with the same GCD are predicted the same, we evaluate model performance by the number of GCD below $100$ predicted correctly, measured on a random test sample of $100,000$ pairs: $1000$ pairs for each GCD from $1$ to $100$. Charton reports a best performance of $22$ correct GCD for a model trained on uniformly distributed inputs. \\
\textbf{Note.} We prefer this test metric over a more standard accuracy on random input pairs, because the GCD are distributed according to an inverse square law. In particular the probability that a GCD is $1$ is about $62\%$. As a result, the accuracy metric results in overly optimistic model performances.

\textbf{Modular multiplication.} 
Modular arithmetic plays an important role in many public key cryptography algorithms~\citep{diffie1976new,Reg05}, and is known to be a hard problem for neural networks~\citep{palamasinvestigating}. Modular addition was studied in several previous works, in the context of grokking~\citep{Power2022Grokking,liu2022towardsgrok} and mechanistic interpretability \citep{zhong2023pizza}\footnote{\cite{Power2022Grokking} also study modular {\em division}, equivalent to modular multiplication.}.  While modular multiplication over $\mathbb{Z}/p\mathbb Z ^\times$ is {\em mathematically} is equivalent to modular addition mod $p-1$, these problems differ {\em computationally}, due to the hardness of the discrete logarithm \citep{diffie1976new}.
 In most previous works on arithmetic modulo $p$, model inputs are sampled from integers between $0$ and $p$, which results in a very small problem space for small $p$. In this work, we study the multiplication modulo $67$ of two integers from $1$ to $1$ million. This allows for a much larger problem space, and training sets. Model accuracy is evaluated by the percentage of correct predictions of $a\times b \mod 67$, on a test set of $10,000$ examples (a new test set is generated at every evaluation). In this problem, all outcomes from $1$ to $66$ are uniformly distributed, while $0$ appears nearly twice as often.
 
\textbf{Eigenvalue calculation.}
This problem was introduced to deep learning by \citet{charton2022linear}, who showed that transformers can learn to predict the eigenvalues of real symmetric matrices with independent and identically distributed entries, rounded to three significant digits. The eigenvalue problem is arguably a  harder problem than the previous two, non-linear and typically solved by iterative algorithms. Note also that because matrix entries and eigenvalues are rounded, this problem features \emph{noisy} inputs and outputs. Model accuracy is evaluated as the percentage of model predictions that predict the correct eigenvalues of a test matrix with less than $5\%$ relative error (in $\ell^1$ distance). It is measured on a test set of $10,000$ samples, generated afresh at every evaluation.

\textbf{Models and tokenizers.}
In all experiments, we use sequence-to-sequence transformers~\citep{vaswani2017attention} with $4$ layers in the encoder and decoder ($4$-layers encoders and $1$-layer decoder for eigenvalues), an embedding dimension of $512$, and $8$ attention heads. Models have $35$ million parameters for GCD and modular multiplication, and $22$ million for eigenvalues. They are trained to minimize a cross-entropy loss, using the Adam optimizer~\citep{kingma2014adam}, with a learning rate of $10^{-5}$, over batches of $64$. The integer inputs and outputs of the GCD and multiplication problems are tokenized as sequences of digits in base $1000$, preceded by a separator token. The real numbers in the eigenvalue problem are encoded as floating point numbers, rounded to three significant digits, and tokenized as the triplet $(s,m,e)$ -- sign, (base $1000$) mantissa, (base $10$) exponent -- i.e. $f = s \cdot m \cdot 10^e$ (P1000 encoding from~\citet{charton2022linear}). All experiments are run on one NVIDIA V100 GPU with $32$ GB of memory. 

\section{Repetition Helps}\label{sec:repetition}

We now embark on a systematic study of the impact of data budget on performance, for various training budgets. In other words, we compare the performance of models trained on datasets with a fixed number of examples (data budget), for increasing amounts of time (training budget).

\begin{figure}[h]
    \center
  \includegraphics[width=0.9\linewidth]{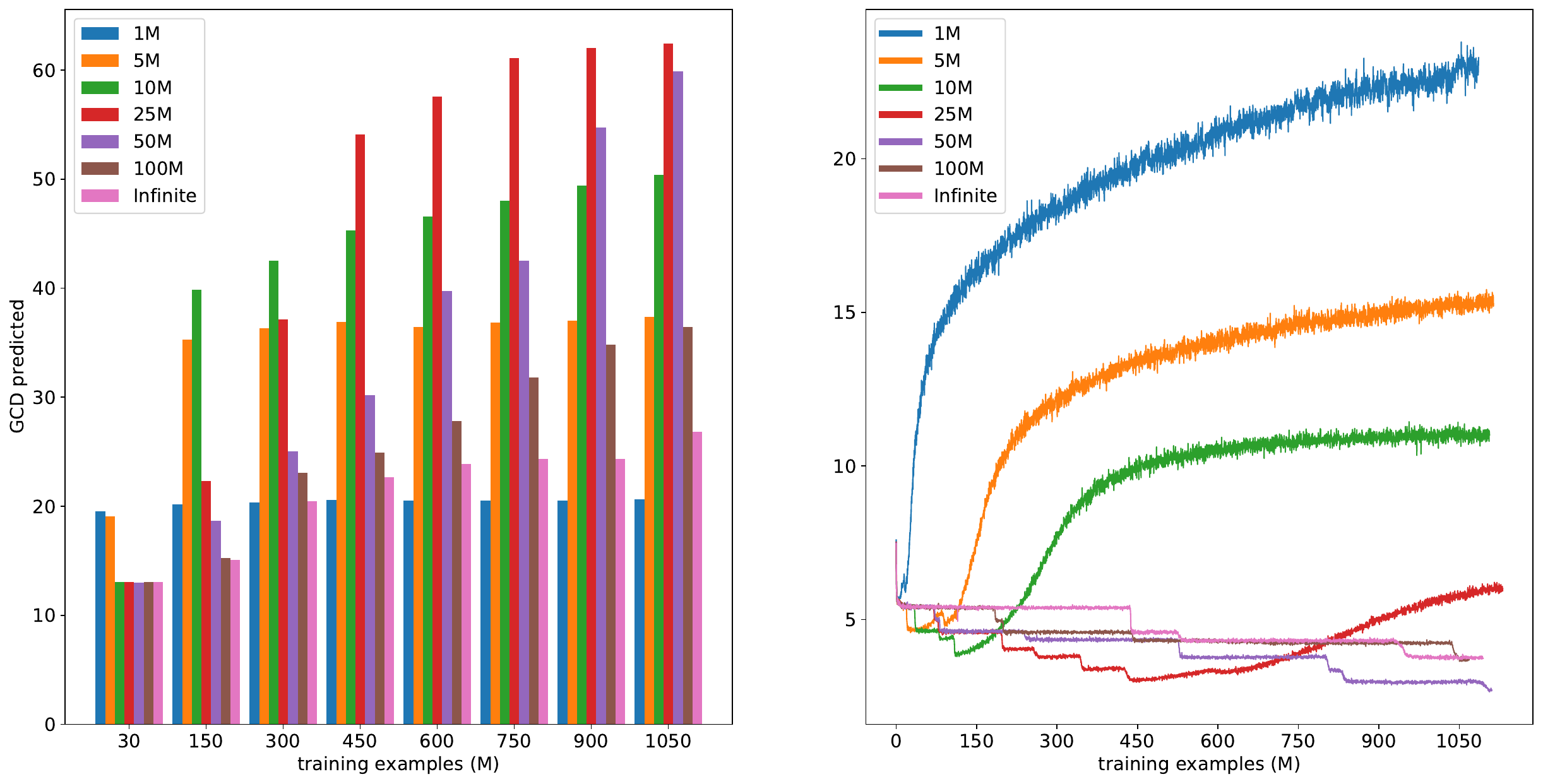}
  \caption{\small{{\bf GCD problem:} (Left) GCD accuracy for different data and training budgets (average of 5 models). (Right) Test loss of models as a function of training budget, for fixed data budgets.}}
  \label{fig:baseline}
\end{figure}

On the {\bf GCD problem,} we consider data budgets of $1, 5, 10, 25, 50$ and $100$M distinct examples, and an ``unlimited data'' setting, where new examples are generated on the fly and DB$\approx$ TB\footnote{For GCD and modular multiplication, input pairs are uniformly sampled integers from $1$ to 1 million. In the unlimited data case, this gives rise to infrequent repetitions: over $\sim 1$ billion input pairs, our largest data budget, no elements are repeated $3$ or more times, and about $500$ thousand are repeated twice.}. 
For each data budget, we train $5$ models with a training budget of over $1$ billion examples, and report their average performance (number of correctly predicted GCD), as the TB increases (Figure~\ref{fig:baseline} Left).

For a modest training budget of $30$ million, the models with the smallest DB ($1$ and $5$ million, $1$M and $5$M-models henceforth) achieve the best performance ($20$ GCD vs $13$ for all other DB). As TB increases, the $1$M-models start overfitting,
as shown by the increasing test losses in Figure~\ref{fig:baseline} (Right), and their performance saturates at $21$ correct GCD. 
The performance of the $5$M models keeps improving to $36$ GCD, for a TB of $150$ million examples, then saturates around $38$ GCD as the models overfit.
For TB of $150$ and $300$ million examples, the best performing models are the $10$M. As training proceeds, they are 
outperformed by the $25$M models, which achieve the best performance for TB from $450$ million to $1.05$ billion examples (with the $50$M-model a close second at $1$ billion). 
Throughout training, the models trained on small data budgets learn faster. However, past a certain TB, they overfit their training data, and their performance saturates.

\textbf{Note.} {\em Overfitting} is an overloaded term. In this paper, we define it by its empirical consequences: {\em a model overfits when its test loss starts increasing, while the train loss continues to decrease}. The relation between learning and overfitting is further studied in Appendix~\ref{app:overfit}.

Conversely, models trained with large or unlimited DB perform the worst. For a TB of one billion examples, the $25$M-models predict $62$ GCD on average, and the $50$M-models $60$. The $100$M-models only predict $37$ GCD and models trained on an unlimited data budget, where all training examples are seen only once, predict $27$ GCD, \emph{way worse} than models trained on $25$M distinct examples, repeated $42$ times on average. 
Summarizing, \textbf{smaller data budgets and more frequent repetition allow for faster learning, but also for much better performance.}

We observe a similar behavior for {\bf modular multiplication}. For a TB of $600$ million, we train $5$ models for small DB, and 25 or 30 for larger DB, to zoom on this interesting region (Table \ref{tab:modmul_base}). Models trained on an unlimited data budget perform at ``chance level'': they always predict $0$ and achieve about $3\%$ accuracy. Models trained on data budgets of $100$ million examples fare little better, and models trained on $10$ million examples or less overfit and do not learn. 

Models trained on DB of $25$M and $50$M (for an average repetition of $24$ and $12$) achieve $40$ and $60\%$ accuracy and exhibit a different behavior. On this task, learning happens in sudden steps, separated by flat plateaus (see the empirical learning curves in Figure \ref{fig:stepmodmul} in Appendix \ref{app:add_exp}), the two last plateaus corresponding to $51\%$ and $99\%$ accuracy. About $25\%$ ($7$ out of $25$) of $50$M models achieve $99\%$ accuracy (i.e. fully learn the task), and almost $90\%$ ($22$/$25$) achieve $50\%$ (i.e. one learning step away). %For $25$M model, the percentages are about $25\%$ and $50\%$. 
On this task, \textbf{learning emerges through repetition}. Models trained on smaller data budgets can perform tasks that models trained from large or unlimited data budget cannot learn. To probe whether learning eventually occurs, we trained $19$ models with unlimited and $100$M DB on increased TB of $2$ billion examples. None of the unlimited DB models could learn modular multiplication, but one $100$M model out of $19$ achieved $99\%$ accuracy after $1$B TB, and $5/19$ after $2$B. 

\begin{table}[h]

\centering
\small
\begin{tabular}{lccccccc}
\toprule

& \multicolumn{7}{c}{Data budget (millions)}\\
 & 1 & 5 & 10 & 25 & 50 & 100 & unlimited\\
\midrule
Average accuracy ($\%$) & 1.6 & 3.8 &4.4 & 40.4 & \textbf{59.5} & 5.4 & 3.0 \\
Number of models achieving 99\% accuracy & 0/5 & 0/5 & 0/5 & 6/25 & \textbf{7/25} & 0/30 & 0/30 \\
Number of models achieving 50\%+ accuracy  & 0/5 & 0/5 & 0/5 & 13/25 & \textbf{22/25} & 0/30 & 0/30\\
Number of models trained & 5 & 5 & 5 & 25 & 25 & 30 & 30 \\ 
\bottomrule
\end{tabular}
\caption{\small \textbf{Multiplication modulo 67}. Accuracy of models trained on a budget of 600 million data points.}
\label{tab:modmul_base}
\end{table}
Finally, on the {\bf eigenvalue problem}, \citet{charton2022linear} trained models with unlimited data budgets (DB$\approx$TB) and
observed that whereas $4$-layer transformers  can learn to compute the eigenvalues of $5\times 5$ matrices, deeper models are required for larger problems: 6-layers for $8\times 8$ matrices, $8$ for $10\times 10$ and $12$ layers for $12\times 12$ matrices. Even with large training budgets, $4$-layer models where unable to learn the eigenvalues of $10$ or $12$ dimensional matrices.

In our experiments, we wanted to study whether smaller DB could \emph{induce} small models to learn large problems. 
We trained {\em $4$-layer} transformers to predict the eigenvalues of $10 \times 10$ matrices. We trained $5$ models for each data budget of $1,5,10,25,50$ and $100$M, and $5$ for an unlimited DB (one pass over the training data), with TB up to $500$ million. 
As expected, none of the models trained on unlimited DB did learn: all test accuracy remained close to $0$. However, $4$ of the $30$ models trained on smaller DB achieved $99\%$ accuracy: $3$ models trained on $50$ million examples (repeated $10$ times), and one model trained on $10$ million (repeated $50$ times). Scaling even further, to $12 \times 12$ matrices, still using $4$-layer transformers, with a TB of 420 millions,  $2$ models (out of $35$) begin learning: a $10$M model achieved $21\%$ accuracy, and a $5$M $3.5\%$.
As in previous experiments, for a given training budget, smaller data budgets and repeated training examples prove beneficial, but on this task, \textbf{small datasets improve model scaling}. With small DB, problems that required $8$-layer or $12$-layer transformers can be learned by $4$-layer models. 

This first series of experiments clearly indicates that \textbf{repetition helps learning.} On three different tasks, for a fixed training budget, models trained on a small data budget, i.e. fewer distinct examples repeated several times, achieve much better performance than models trained from examples used only once or repeated a small number of times, as is customary in most recent works on language models \citep{muennighoff2023fourepochs}.

This phenomenon applies in different ways for different problems. On the GCD task, small DB allow for faster learning and higher accuracy. For modular multiplication, we observe emergence: a task inaccessible to models trained with large or unlimited DB is learned with small DB. Finally, for eigenvalues, small DB allow for better model scaling: tasks that normally require $8$ or $12$-layer transformers are learned by $4$-layer models.
But in all cases, the repetition achieved by small DB prove beneficial:
{\bf smaller data budgets with repetition can elicit ``emergent learning''}.

\section{Two-set training}\label{sec:2-set}

The previous experiments demonstrate that for a fixed training budget, the optimal data budget is not the largest possible, as commonly practiced. On all three tasks, training from a set of distinct examples an order of magnitude smaller than the training budget, repeated many times, improves performance. We now turn to a different but related problem: how to best use a given data budget?

As we have seen, repeated examples help the model learn. Training from a small subset of the available data should therefore be beneficial, since it would increase repetition. However, models trained from very small datasets will eventually overfit their data, causing their accuracy to saturate. Yet, this can be prevented by increasing the size of the training set. To address these contradictory requirements -- a small train set to increase repetition vs a large train set to avoid overfitting -- we propose \emph{two-set training}. We randomly split the training sample into a small set of examples that will be repeated many times during training, and a large set of examples that will be seen a few times only. By doing so, we hope that the small set fosters learning, while the large set prevents overfit.

Specifically, for a data budget of $N$ distinct examples, we randomly select $S<N$ examples that will form the repeated set -- in practice, we shuffle the training set, and assign the $S$ first examples to the repeated set. During training, examples are selected from the repeated set with probability $p$, and from the $N-S$ others with probability $(1-p)$.
As a result, a model trained with a training budget of $T$ will see $pT$ examples from the repeated set, repeated $pT/S$ times on average, while the $N-S$ remaining examples will be repeated $(1-p)T/(N-S)$ times on average. 
The repetition levels in both samples can be adjusted by choosing the values of $S$ and $p$. Note that the limiting cases $p=0$ and $p=1$ correspond to one-set training, with a data budget of $N-S$ and $S$ examples respectively.

On the \textbf{GCD problem}, models trained on a single set, with a data budget of $100$ million examples and a training budget of $600$ million, predict $27$ GCD on average (Figure \ref{fig:baseline} (Left)). Experimenting with two-set training for different values of $S$ and $p$, we observe that models trained on a repeated set of $250,000$ examples or less, with a probability $p$ of $0.25$ or $0.5$, predict more than $62$ GCD on average, a much better performance than their one-set counterparts. For $S=50,000$ and $p=0.25$, models predict $69$ GCD on average, a better performance than the best models trained on a single set, with a larger training budget of $1$ billion examples. For these parameters, the $50$k examples in the small set are repeated $3,000$ times on average, and the rest of the training examples $4.5$ times on average. On a $100$M data budget, two-set training clearly outperforms single set training.

\begin{figure}[t]
\center\small
\includegraphics[width=0.8\textwidth]{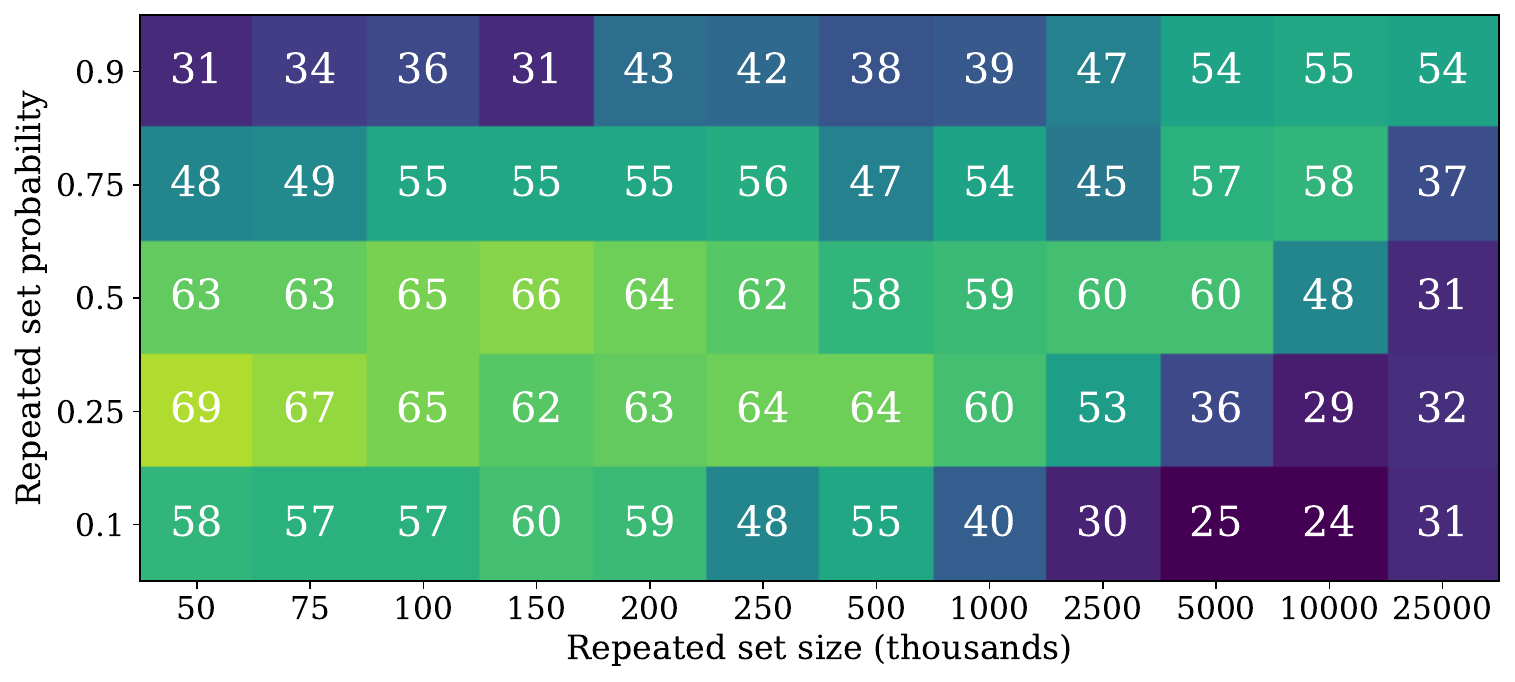}
  \caption{\small {\textbf{Two-set training for the GCD problem:} Number of correctly predicted GCD as a function of $S$ and $p$. Each measurement is the average of $6$ models. Data budget $100$M, training budget $600$M. Note the high performance for very small sets $S$ of sizes $50$, $75$, $100$, $150$ and $200$ thousand, with $p=0.25$ and $p=0.5$.}}
  \label{fig:twosamples}
  \vspace{-0.45cm}
\end{figure}

These results can be extended to unlimited training sets, by creating a fixed set of $S$ examples, selected with probability $p$, and generating (unlimited) random examples with probability $1-p$. The best choices of $p$ and $S$ are roughly the same as with a DB of $100$M (Figure~\ref{fig:twosamples_inf} in Appendix~\ref{app:add_exp}). In particular, with $p=0.25$ and $S=50,000$, two-set training on unlimited data achieves an average performance of $67$ GCD on $6$ models, a spectacular improvement over models trained on unlimited (single) datasets, which predict $25$ GCD on average.

Therefore, for large and unlimited data budgets, frequent repetition of a tiny number of random examples, lost in a sea of single-use examples, unlocks surprising performance gains. Note the synergistic nature of this effect: training on the tiny sample alone (with large repetition), or one-set training on the same data budget, result in much lower performance than what two-set training provides by mixing them together (see also Appendix~\ref{app:monobatch}: during training, mixing repeated and single-use examples into the same mini-batches is required for two-set training to happen).

\begin{figure}[h]
\vspace{-0.15cm}
\includegraphics[width=\textwidth]{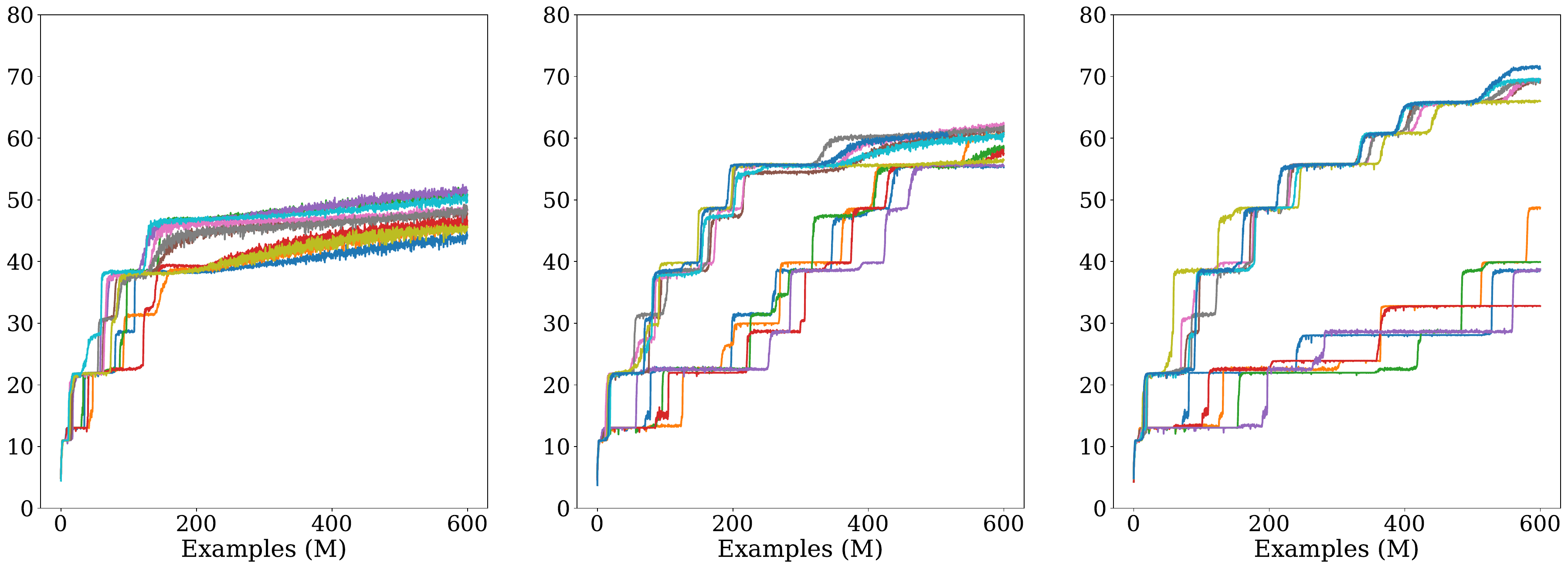}
  \caption{\small {\textbf{Two-set versus single-set training for the GCD problem:} Number of correct GCD as a function of training budget(up to $600$M) for data budgets of $10$M (left), $25$M (center), and $50$M (right). Two-set training with $p=0.25$ and $S=50,000$ (top 6 curves)  versus single-set training (lower $6$ curves). See Figure \ref{fig:twosamples_25M2} in Appendix \ref{app:add_exp} for extended TB with DB of $50$M. }}
  \label{fig:twosamples_smaller}
  
\end{figure}

We observe similar behavior for smaller data budgets. Figure~\ref{fig:twosamples_smaller} compares single and two-set training performance, for data budgets of $10$, $25$ and $50$ million example, and training budgets up to $600$M. For a given training budget, two-set training always achieves better performance than single-set training, and the benefit of two-set training increases as DB get larger. On this problem, two-set training accelerates learning. With large enough TB, single-set models sometimes catch up with the performance of their two-set counterparts with large enough TB (for $400$M TB for $10$M models, $600$M for $25$M, $1$B TB for $50$M models, see Figure~\ref{fig:twosamples_25M2}, Appendix~\ref{app:add_exp}). Still, most two-set models retain a marginal advantage\footnote{and note that $S$ and $p$ might no longer be optimal for this larger training budget} over models trained on a single set.

\begin{figure}[t]
\begin{minipage}{0.5\textwidth} 
\includegraphics[width=\textwidth]{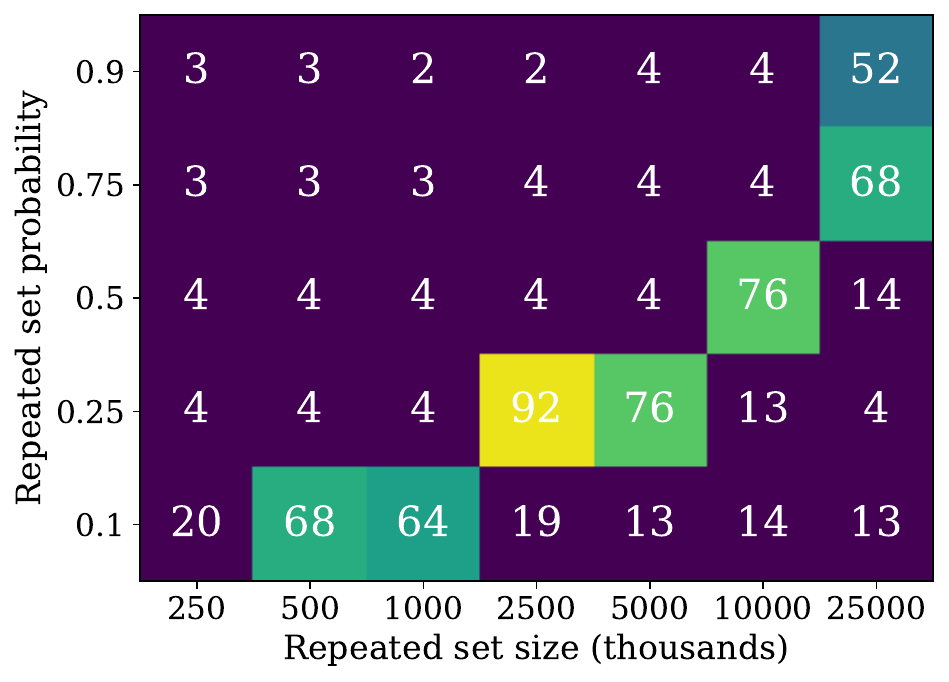}
\end{minipage}\hfill 
\begin{minipage}{0.5\textwidth}
  \includegraphics[width=\textwidth]{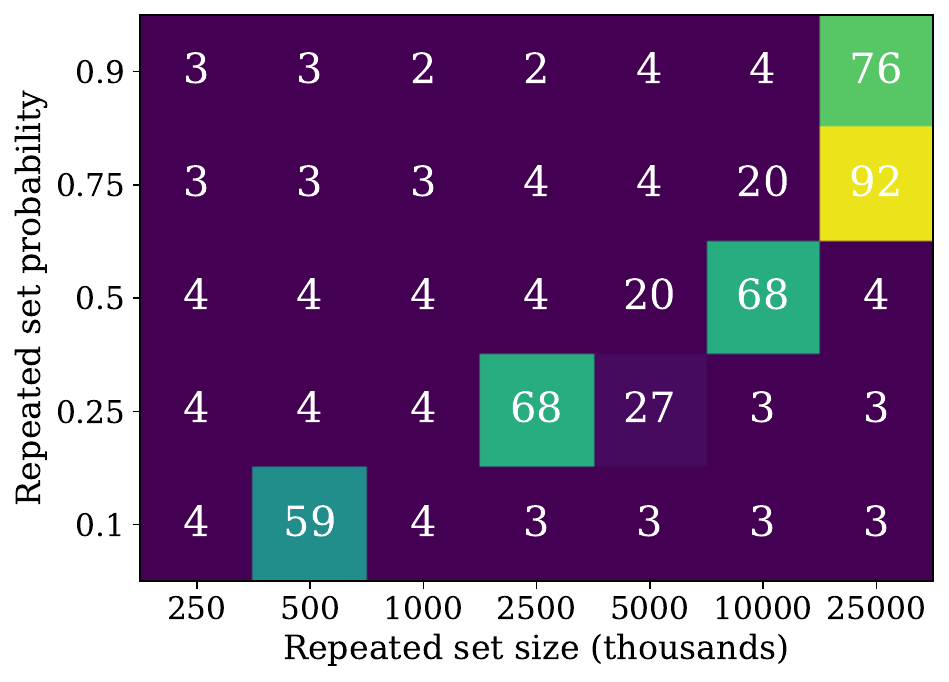}
\end{minipage}
  \caption{\small{{\bf Two-set training for Modular Multiplication:} Accuracy as a function of small set size $S$ and $p$, each averaged over $6$ models. Data budget $100$M (left) and unlimited (right), training budget $600$M. Note: the bottom right of the left graph correspond to single-set $10$M-models: for $p=0.1$ and $S=10$M, the small and large set are selected with the same probability.}}
  \label{fig:twosamples_mod67_1}
%  \vspace{-10pt}
\end{figure}

For \textbf{modular multiplication}, experiments with large and infinite data budget, for a training budget of $600$M (Figure~\ref{fig:twosamples_mod67_1}), indicate that larger repeated samples and smaller repetition, are needed, compared to GCD. With a DB of $100$M, $S$ should be selected between $2.5$ and $10$ million examples, and $p$ between $0.25$ and $0.5$, for a small set repetition between $30$ and $60$ (vs $3000$ for the GCD experiments). For unlimited DB, $S=25$M and $0.75\leq p \leq 0.9$, a repetition between $18$ and $22$, seem optimal. Note also that in this problem, the choice of parameters $S$ and $p$ is more sentitive: only a few combinations allow for good performance (empirically,  constant ratio between repetition on the small and large sample ($\frac{p(N-S)}{(1-p)S}\approx 10$).

However, with a careful choice of $p$ and $S$, two-set training achieves better performance than single set training for all data budgets from $25$M to unlimited. Table~\ref{tab:modmul_2samples_0} presents the proportion of models, trained on single and two sets, that learn to compute multiplication modulo $67$, after a training budget of $600$M. With two set training, $50$ to $58\%$ of the models learn multiplication with $99\%$ accuracy. With single set training, $24$ to $28\%$ learn for DB $25$ and $50$M, and none for larger DB. In these experiments, two-set training improves accuracy for all data budgets. However, its impact on learning speed (observed for GCD) is less conclusive (Table~\ref{tab:modmul_2samples_app} in Appendix \ref{app:add_exp}).

\begin{table}[h]
\small
\centering
\begin{tabular}{lc|cc|cc}
\toprule
& &\multicolumn{2}{c|}{Two sets} & \multicolumn{2}{c}{Single set}\\
Data budget & p / S & $>50\%$ & $>99\%$  & $>50\%$ & $>99\%$\\
\midrule
25M& 0.1 / 1M & 50 & \textbf{50} & 52 & 24\\
50M & 0.25 / 2.5M & 90 & \textbf{50} & 88 & 28 \\
100M &  0.5 / 10M & 88 & \textbf{54} & 0 & 0 \\
Unlimited & 0.25 / 2.5M & 92 & \textbf{58} & 0 & 0 \\
\bottomrule
\end{tabular}
\caption{\small {\bf Two-set training on modular multiplication.} Percentage of models (different random initializations) learning to compute modular multiplication with $50$ and $99\%$ accuracy. Training budget: $600$M. For DB $25$M and $50$M, $10$ models with two-set training, and $25$ with single set training. For DB $100$M and unlimited, $26$ models with two-set training, and $30$ with single set training. }
\label{tab:modmul_2samples_0}
\end{table}

Finally, on the \textbf{eigenvalue problem} for $10\times 10$ matrices, we train models with an unlimited data budget and a training budget of $500$M. With these parameters, models trained on single sets do not learn (see Section \ref{sec:repetition}), but two-set training achieves significant accuracy. For $p=0.25$ we run $15$ models each for $7$ different sizes of
 $S$ between $15,000$ and $960,000$. $9$ models out of $105$ learn to predict with more than $60\%$ accuracy %. As in the case of modular multiplication, large repeated sets improve model performance 
(Table~\ref{tab:eigen_2samples_0}). We see that injecting small, frequently repeated random subsets into the training data causes emergence of learning, where uniform repetition fails! Note, again, the synergistic effect: neither training on the small set alone, nor training with unlimited data budget in one epoch would allow any learning at all - it is the combination of both that makes two-set training powerful!

\begin{table}[h]
\small
\centering
\begin{tabular}{l|ccccccc|c}
\toprule
 &\multicolumn{7}{c|}{Two sets} & Single set\\
S (thousands) & 960 & 480 & 240 & 120 & 60 & 30 & 15 &\\
\midrule
\#models learning with 60\% accuracy & 3/15 & 1/15 & 1/15 & 3/15 & 0/15 & 0/15 & 1/15 & 0/5\\
\bottomrule
\end{tabular}
\caption{\small {\bf Two-set training on eigenvalues.} Number of models (different random initializations) learning to compute eigenvalues with over $60\%$ accuracy. Training budget: $500$M. Larger small sets achieve better results, single-set learning does not learn. }
\label{tab:eigen_2samples_0}
\end{table}

Overall, our experiments indicate that, for a given data budget, two-set training -- repeating a small set of \emph{randomly selected} during training -- greatly improves model performance, either by accelerating learning (GCD), or increasing model accuracy (modular multiplication,
 eigenvalues). The size of the repeated set appears to be problem dependent: small for GCD, larger for modular multiplication and eigenvalues. 

\section{Ablations and variations}\label{sec:ablations}

In this section, we discuss possible improvements to two-set training. Detailed ablation results can be found in Appendix\ref{app:ablation}.

\paragraph{Curating the repeated sample.} In two-set training, repeated examples are randomly sampled from the available training data. We now experiment with a possible improvement: selecting the repeated examples. Perhaps what really matters is the repetition of a particular class of ``informative'' examples, as in curriculum learning. 
The GCD problem is particularly well suited for this type of investigation. \citet{charton2023gcd} showed that increasing the proportion of small integers, or oversampling the tails of the distribution of GCD in the training set ($\text{Prob}(\text{GCD}=k)\sim \frac{1}{k^2}$), greatly improved model performance.

We experimented with three curation strategies for the repeated set: log-uniform and uniform distributions of operands and input, shown to be beneficial by Charton, ``easy sets'' featuring small input and outcomes, and ``heavy tail sets'' featuring large GCD. For each setting, we trained $5$ models with four ``good choices'' of $S$ and $p$ (Table~\ref{tab:gcd_xp3_0}), a data budget of $100$M and training budget of $600$M.

\begin{table} [h]
\centering
\small
\begin{tabular}{lcccc}
\toprule
 S / p & 50k / 0.25 & 150k / 0.25 & 150k / 0.5 & 500K / 0.5\\
\midrule
Log-uniform inputs & 55.9 & 59.4 & 57.9 & 62.0 \\
Uniform GCD & 55.9 & 54.5 & 41.9 & 54.9  \\
Log-uniform inputs and GCD & 62.2 & {\bf71.7} & {\bf 66.5} & {\bf 72.6} \\
\midrule
Small inputs (1-1000) & 61.2 & {\bf 67.5} & 62.6 & 62.9 \\
GCD 1- 10 & 59.9 & 63.8 & 55.8 & 62.3 \\
GCD products of 2 and 5 & 54.2 & 39.8 & 40.7 & 30.1 \\
\midrule
All GCD but 1 & {\bf 65.4} & 63.7 & 56.7 & 58.1\\
%All GCD but 1,2 & {\bf 66.8} & 60.0 & 62.8 & 56.9\\
All GCD but 1,2,3 & {\bf 66.7} & 58.4 & 62.8 & 58.2  \\
%All GCD but 1,2,3,4 & {\bf 65.5} & 60.3 & 62.8 & 56.9 \\
All GCD but 1,2,3,4,5 & {\bf 66.5} & 60.6 & 64.9 & 56.3 \\
\midrule
Baseline (two-set training from random examples) & {\bf 69.4} & 61.9 & {\bf 65.9} & 59.4 \\
\bottomrule
\end{tabular}
\caption{\small \textbf{GCD problem: cherry-picking the repeated set}. Number of GCD predicted, average of 5 models (3 for baseline), training budget 600M. \textbf{bold}: more than 65 GCD predicted.}
\label{tab:gcd_xp3_0}
\vspace{-8pt}
\end{table}

These strategies do not achieve better results than the baseline two-set training with a random repeated set. A slight improvement is observed when repeated samples are selected from a log-uniform input and GCD (for which \citet{charton2023gcd} reports $91$ correct GCD for single-set training). Overall, we find  that repeated set curation has, at best, a marginal impact on performance.  This is a counter-intuitive but significant result.

\textbf{Shifting the repeated sample.} In the GCD experiments, with $p=0.25$ and $S=50,000$, repeated examples are seen $3000$ times for a training budget of $600$M. Since this large repetition may lead to overfit, we experimented with ``shifting samples'': replacing the repeated examples after a $k$ repetitions. In Appendix~\ref{app:shift}, we experiment with $k$ from $10$ to $100$, and observe that this has no impact on model performance.

\textbf{Batching matters.} All models in this paper are trained on mini-batches of $64$ examples. In two-set training, batches mix examples from the repeated and the large set. We experimented with batches that only use samples from one set at a time. For instance, when training with $p=0.25$, $25\%$ of batches would use repeated examples only. For both GCD and modular multiplication, we observe that models trained on batches from one sample only fail to learn. This indicates that mixing repeated and non-repeated examples is required for two-set training to happen (see also Appendix~\ref{app:monobatch}).

\textbf{From two to many-set training.} Two-set training effectively makes the training sample non identically-distributed: examples from the repeated sample occur with a larger probability. We can generalize this method by introducing a probability distribution $P$ on the training examples, such that for any $i\leq N$, $P(i)$ is the probability that the $i$-th example is selected during training. In two-set training, $P$ is a step function distribution with two values: $p/S$ and $(1-p)/(N-S)$, we now replace it with a discrete exponential distribution $P(i) \sim \beta e^{-\beta i / N},$ with $\beta > 0$, suitably normalized. Table~\ref{tab:gcd_xp4_main} presents the performance of models trained on the GCD problem with such ``continuous'' data distributions, indicating that our observations on two-set training generalize to such data sampling techniques. More details, and results on modular multiplication, can be found in Appendix~\ref{app:manyset}. These results suggests that our observations on two-set training can be extended to a wider class of methods, that use non-uniform sampling over a randomly ordered training set. 
\begin{table} [h]
\vspace{-5pt}
\centering
\small
\begin{tabular}{lccccccccccccc}
\toprule
$S_{\text{eff}}$ & 25k & 50k & 100k& 250k & 500k & 1M & 1.5M &  2M & 2.5M & 3M & 3.5M & 4M & 5M \\
$\beta$ & 1152 & 576 & 288 & 115 & 58 & 29 & 19 & 14& 11.5 & 9.6 & 8.2 & 7.2 &  5.8\\
\midrule
GCD &19 & 21 & 29 & 38 & 46 & 55 & 56 & 57 & 61 & 65 & 63 & 62 &56 \\
\bottomrule
\end{tabular}
\caption{\small \textbf{GCD for different exponential distributions.} Correctly predicted GCD, best of 5 models, trained on 600 million examples.}
\label{tab:gcd_xp4_main}
\vspace{-17pt}
\end{table}

\section{Discussion}

Our findings indicate that repetition, and possibly memorization, fosters learning. They suggest that models should be trained on datasets of repeated, but not necessarily curated examples, and that amplifying a randomly chosen subset of the training data may bring additional learning benefits. Two-set training is easy to implement, and applicable to a large variety of situations. Its extension to smooth distributions allows for finer control over repetition levels in the training sets.

We can contemplate how our observations carry over to large language models (LLM) trained on natural data. An important factor is the presence of repetition in the training data. We believe that pre-training corpora -- text scraped from the internet, public code repositories -- feature many repeated examples (quotes, copied passages, duplicated functions), and that the phenomena we describe are already at work in LLMs during the pre-training stage. Fine-tuning corpora, on the other hand, are often curated and feature less repetition. We believe two-set training, and associated methods, may prove beneficial for fine-tuning LLMs.

Our observations on two-set training are thought-provoking and deserve further study. The fact that the repeated set can be chosen at random, and that curating repeated examples bring little to no improvement in performance suggest that what matters, here, is seeing the \emph{exact same} example several times. The particulars of the example, its informational value, interest, whether it is typical or exceptional, seem to have little impact.  This is all the more curious as, even in the two-set setting, repetition occurs at a very low frequency. In the two-set GCD experiments, repeated examples were seen $3000$ times over a training budget of $600$ million: once every 200,000 examples on average. The frequency is even lower for modular multiplication. Besides, the repeated examples are mixed with non-repeated examples into mini-batches, and our experiments indicate that this mixing is required for the two-set effect to appear. Still, this very infrequent repetition, and mini-batch mixing, brings a significant boost in model performance. 

This raises several tantalizing questions: how does the transformer ``figure'' that a given example, lost in a mini-batch, has been seen, several hundred thousand examples before? Our research suggests that there exists a qualitative difference between ``déjà vu'' and ``jamais vu'' examples -- data points the model has already seen, or never seen. How do transformers, and perhaps other architectures, identify, and then process, ``déjà vu'' examples? To our knowledge, this aspect was overlooked in many prior works on model interpretation. 
We believe our findings point to a number of interesting questions about memorization in language models. This is an intriguing subject for further study.

\newpage
\bibliography{lowdata}

\begin{thebibliography}{41}
\providecommand{\natexlab}[1]{#1}
\providecommand{\url}[1]{\texttt{#1}}
\expandafter\ifx\csname urlstyle\endcsname\relax
  \providecommand{\doi}[1]{doi: #1}\else
  \providecommand{\doi}{doi: \begingroup \urlstyle{rm}\Url}\fi

\bibitem[Allen-Zhu \& Li(2024)Allen-Zhu and Li]{allen-zhu2024physics}
Zeyuan Allen-Zhu and Yuanzhi Li.
\newblock Physics of language models: Part 3.1, knowledge storage and extraction.
\newblock In \emph{Forty-first International Conference on Machine Learning}, 2024.
\newblock URL \url{https://openreview.net/forum?id=5x788rqbcj}.

\bibitem[Ambridge et~al.(2015)Ambridge, Kidd, Rowland, and Theakston]{CogScirepetitio2015}
Ben Ambridge, Evan Kidd, Caroline~F. Rowland, and Anna~L. Theakston.
\newblock The ubiquity of frequency effects in first language acquisition.
\newblock \emph{J Child Lang}, 42\penalty0 (2):\penalty0 239--273, 2015.

\bibitem[Bartlett et~al.(2020)Bartlett, Long, Lugosi, and Tsigler]{BenignPNAS2020}
Peter~L. Bartlett, Philip~M. Long, Gábor Lugosi, and Alexander Tsigler.
\newblock Benign overfitting in linear regression.
\newblock \emph{Proceedings of the National Academy of Sciences}, 117\penalty0 (48):\penalty0 30063--30070, 2020.
\newblock \doi{10.1073/pnas.1907378117}.
\newblock URL \url{https://www.pnas.org/doi/abs/10.1073/pnas.1907378117}.

\bibitem[Bartlett et~al.(2021)Bartlett, Montanari, and Rakhlin]{Bartlett_Montanari_Rakhlin_2021benign}
Peter~L. Bartlett, Andrea Montanari, and Alexander Rakhlin.
\newblock Deep learning: a statistical viewpoint.
\newblock \emph{Acta Numerica}, 30:\penalty0 87–201, 2021.

\bibitem[Belkin(2021)]{Belkin_2021benign}
Mikhail Belkin.
\newblock Fit without fear: remarkable mathematical phenomena of deep learning through the prism of interpolation.
\newblock \emph{Acta Numerica}, 30:\penalty0 203–248, 2021.

\bibitem[Bengio et~al.(2009)Bengio, Louradour, Collobert, and Weston]{BengioCurriculum2009}
Yoshua Bengio, J\'{e}r\^{o}me Louradour, Ronan Collobert, and Jason Weston.
\newblock Curriculum learning.
\newblock In \emph{Proceedings of the 26th Annual International Conference on Machine Learning}, ICML '09, pp.\  41–48, New York, NY, USA, 2009. Association for Computing Machinery.
\newblock URL \url{https://doi.org/10.1145/1553374.1553380}.

\bibitem[Brown et~al.(2020)Brown, Mann, Ryder, Subbiah, Kaplan, Dhariwal, Neelakantan, Shyam, Sastry, Askell, Agarwal, Herbert-Voss, Krueger, Henighan, Child, Ramesh, Ziegler, Wu, Winter, Hesse, Chen, Sigler, Litwin, Gray, Chess, Clark, Berner, McCandlish, Radford, Sutskever, and Amodei]{GPT3brown2020}
Tom~B. Brown, Benjamin Mann, Nick Ryder, Melanie Subbiah, Jared Kaplan, Prafulla Dhariwal, Arvind Neelakantan, Pranav Shyam, Girish Sastry, Amanda Askell, Sandhini Agarwal, Ariel Herbert-Voss, Gretchen Krueger, Tom Henighan, Rewon Child, Aditya Ramesh, Daniel~M. Ziegler, Jeffrey Wu, Clemens Winter, Christopher Hesse, Mark Chen, Eric Sigler, Mateusz Litwin, Scott Gray, Benjamin Chess, Jack Clark, Christopher Berner, Sam McCandlish, Alec Radford, Ilya Sutskever, and Dario Amodei.
\newblock Language models are few-shot learners.
\newblock \emph{arXiv preprint arXiv:2005.14165}, 2020.

\bibitem[Charton(2022)]{charton2022linear}
Fran\c{c}ois Charton.
\newblock Linear algebra with transformers.
\newblock \emph{Transactions on Machine Learning Research}, 2022.
\newblock ISSN 2835-8856.
\newblock URL \url{https://openreview.net/forum?id=Hp4g7FAXXG}.

\bibitem[Charton(2024)]{charton2023gcd}
Fran{\c{c}}ois Charton.
\newblock Learning the greatest common divisor: explaining transformer predictions.
\newblock In \emph{The Twelfth International Conference on Learning Representations}, 2024.
\newblock URL \url{https://openreview.net/forum?id=cmcD05NPKa}.

\bibitem[Devlin et~al.(2019)Devlin, Chang, Lee, and Toutanova]{BERT}
Jacob Devlin, Ming{-}Wei Chang, Kenton Lee, and Kristina Toutanova.
\newblock {BERT:} pre-training of deep bidirectional transformers for language understanding.
\newblock In \emph{17th Annual Conference of the North American Chapter of the Association for Computational Linguistics: Human Language Technologies}, 2019.

\bibitem[Diffie \& Hellman(1976)Diffie and Hellman]{diffie1976new}
Whitfield Diffie and Martin Hellman.
\newblock New directions in cryptography.
\newblock \emph{IEEE transactions on Information Theory}, 1976.

\bibitem[Dohmatob et~al.(2024)Dohmatob, Feng, Yang, Charton, and Kempe]{dohmatob2024aTaleTails}
Elvis Dohmatob, Yunzhen Feng, Pu~Yang, Fran{\c{c}}ois Charton, and Julia Kempe.
\newblock A tale of tails: Model collapse as a change of scaling laws.
\newblock In \emph{Forty-first International Conference on Machine Learning}, 2024.
\newblock URL \url{https://openreview.net/forum?id=KVvku47shW}.

\bibitem[Dubey \& et~al.(2024)Dubey and et~al.]{llama3_2024}
Abhimanyu Dubey and et~al.
\newblock The llama 3 herd of models.
\newblock \emph{arXiv preprint arXiv:2407.21783}, 2024.

\bibitem[Feng et~al.(2024)Feng, Dohmatob, Yang, Charton, and Kempe]{feng2024beyond}
Yunzhen Feng, Elvis Dohmatob, Pu~Yang, Fran{\c{c}}ois Charton, and Julia Kempe.
\newblock Beyond model collapse: Scaling up with synthesized data requires reinforcement.
\newblock In \emph{ICML 2024 Workshop on Theoretical Foundations of Foundation Models}, 2024.
\newblock URL \url{https://openreview.net/forum?id=iqoqtNyVta}.

\bibitem[Garg et~al.(2022)Garg, Tsipras, Liang, and Valiant]{NEURIPS2022LiangInContext}
Shivam Garg, Dimitris Tsipras, Percy~S Liang, and Gregory Valiant.
\newblock What can transformers learn in-context? a case study of simple function classes.
\newblock In S.~Koyejo, S.~Mohamed, A.~Agarwal, D.~Belgrave, K.~Cho, and A.~Oh (eds.), \emph{Advances in Neural Information Processing Systems}, volume~35, pp.\  30583--30598. Curran Associates, Inc., 2022.
\newblock URL \url{https://proceedings.neurips.cc/paper_files/paper/2022/file/c529dba08a146ea8d6cf715ae8930cbe-Paper-Conference.pdf}.

\bibitem[Gulrajani \& Lopez-Paz(2021)Gulrajani and Lopez-Paz]{gulrajani2021:ood}
Ishaan Gulrajani and David Lopez-Paz.
\newblock In search of lost domain generalization.
\newblock In \emph{International Conference on Learning Representations}, 2021.
\newblock URL \url{https://openreview.net/forum?id=lQdXeXDoWtI}.

\bibitem[Hernandez et~al.(2022)Hernandez, Brown, Conerly, DasSarma, Drain, El-Showk, Elhage, Hatfield-Dodds, Henighan, Hume, Johnston, Mann, Olah, Olsson, Amodei, Joseph, Kaplan, and McCandlish]{hernandez2022scalinglawsinterpretabilitylearning}
Danny Hernandez, Tom Brown, Tom Conerly, Nova DasSarma, Dawn Drain, Sheer El-Showk, Nelson Elhage, Zac Hatfield-Dodds, Tom Henighan, Tristan Hume, Scott Johnston, Ben Mann, Chris Olah, Catherine Olsson, Dario Amodei, Nicholas Joseph, Jared Kaplan, and Sam McCandlish.
\newblock Scaling laws and interpretability of learning from repeated data, 2022.
\newblock URL \url{https://arxiv.org/abs/2205.10487}.

\bibitem[Hoffmann et~al.(2022)Hoffmann, Borgeaud, Mensch, Buchatskaya, Cai, Rutherford, de~Las~Casas, Hendricks, Welbl, Clark, Hennigan, Noland, Millican, van~den Driessche, Damoc, Guy, Osindero, Simonyan, Elsen, Rae, Vinyals, and Sifre]{chinchillahoffmann2022}
Jordan Hoffmann, Sebastian Borgeaud, Arthur Mensch, Elena Buchatskaya, Trevor Cai, Eliza Rutherford, Diego de~Las~Casas, Lisa~Anne Hendricks, Johannes Welbl, Aidan Clark, Tom Hennigan, Eric Noland, Katie Millican, George van~den Driessche, Bogdan Damoc, Aurelia Guy, Simon Osindero, Karen Simonyan, Erich Elsen, Jack~W. Rae, Oriol Vinyals, and Laurent Sifre.
\newblock Training compute-optimal large language models.
\newblock \emph{arXiv preprint arXiv:2203.15556}, 2022.

\bibitem[Kingma \& Ba(2014)Kingma and Ba]{kingma2014adam}
Diederik~P Kingma and Jimmy Ba.
\newblock Adam: A method for stochastic optimization.
\newblock \emph{arXiv preprint arXiv:1412.6980}, 2014.

\bibitem[Komatsuzaki(2019)]{komatsuzaki2019epochneed}
Aran Komatsuzaki.
\newblock One epoch is all you need.
\newblock \emph{ArXiv preprint, arXiv:1906.06669}, 2019.

\bibitem[Liu et~al.(2022{\natexlab{a}})Liu, Kitouni, Nolte, Michaud, Tegmark, and Williams]{liu2022towardsgrok}
Ziming Liu, Ouail Kitouni, Niklas Nolte, Eric~J Michaud, Max Tegmark, and Mike Williams.
\newblock Towards understanding grokking: An effective theory of representation learning.
\newblock In Alice~H. Oh, Alekh Agarwal, Danielle Belgrave, and Kyunghyun Cho (eds.), \emph{Advances in Neural Information Processing Systems}, 2022{\natexlab{a}}.
\newblock URL \url{https://openreview.net/forum?id=6at6rB3IZm}.

\bibitem[Liu et~al.(2022{\natexlab{b}})Liu, Kitouni, Nolte, Michaud, Tegmark, and Williams]{NEURIPS2022grok}
Ziming Liu, Ouail Kitouni, Niklas~S Nolte, Eric Michaud, Max Tegmark, and Mike Williams.
\newblock Towards understanding grokking: An effective theory of representation learning.
\newblock In S.~Koyejo, S.~Mohamed, A.~Agarwal, D.~Belgrave, K.~Cho, and A.~Oh (eds.), \emph{Advances in Neural Information Processing Systems}, volume~35, pp.\  34651--34663. Curran Associates, Inc., 2022{\natexlab{b}}.
\newblock URL \url{https://proceedings.neurips.cc/paper_files/paper/2022/file/dfc310e81992d2e4cedc09ac47eff13e-Paper-Conference.pdf}.

\bibitem[Liu et~al.(2023)Liu, Michaud, and Tegmark]{liu2023omnigrok}
Ziming Liu, Eric~J Michaud, and Max Tegmark.
\newblock Omnigrok: Grokking beyond algorithmic data.
\newblock In \emph{The Eleventh International Conference on Learning Representations}, 2023.
\newblock URL \url{https://openreview.net/forum?id=zDiHoIWa0q1}.

\bibitem[Lopez-Paz(2025)]{tipLopePaz}
David Lopez-Paz.
\newblock \emph{The Invariance Principle}.
\newblock MIT Press, 2025.

\bibitem[Mohri et~al.(2018)Mohri, Rostamizadeh, and Talwalkar]{MohriLearningBook2018}
Mehryar Mohri, Afshin Rostamizadeh, and Ameet Talwalkar.
\newblock \emph{Foundations of Machine Learning}.
\newblock MIT Press, 2018.

\bibitem[Muennighoff et~al.(2023)Muennighoff, Rush, Barak, Scao, Tazi, Piktus, Pyysalo, Wolf, and Raffel]{muennighoff2023fourepochs}
Niklas Muennighoff, Alexander~M Rush, Boaz Barak, Teven~Le Scao, Nouamane Tazi, Aleksandra Piktus, Sampo Pyysalo, Thomas Wolf, and Colin Raffel.
\newblock Scaling data-constrained language models.
\newblock In \emph{Thirty-seventh Conference on Neural Information Processing Systems}, 2023.
\newblock URL \url{https://openreview.net/forum?id=j5BuTrEj35}.

\bibitem[Palamas(2017)]{palamasinvestigating}
Theodoros Palamas.
\newblock Investigating the ability of neural networks to learn simple modular arithmetic.
\newblock 2017.

\bibitem[Power et~al.(2022)Power, Burda, Edwards, Babuschkin, and Misra]{Power2022Grokking}
Alethea Power, Yuri Burda, Harrison Edwards, Igor Babuschkin, and Vedant Misra.
\newblock Grokking: Generalization beyond overfitting on small algorithmic datasets.
\newblock \emph{ArXiv}, abs/2201.02177, 2022.
\newblock URL \url{https://api.semanticscholar.org/CorpusID:245769834}.

\bibitem[Proust(1919)]{ProustSwann}
Marcel Proust.
\newblock \emph{A la recherche du temps perdu: Du c\^ot\'e de chez Swann}.
\newblock Gallimard, 1919.

\bibitem[Radford \& Narasimhan(2018)Radford and Narasimhan]{Radford2018ImprovingLU}
Alec Radford and Karthik Narasimhan.
\newblock Improving language understanding by generative pre-training, 2018.
\newblock URL \url{https://api.semanticscholar.org/CorpusID:49313245}.

\bibitem[Radford et~al.(2019)Radford, Wu, Child, Luan, Amodei, Sutskever, et~al.]{radford2019language}
Alec Radford, Jeffrey Wu, Rewon Child, David Luan, Dario Amodei, Ilya Sutskever, et~al.
\newblock Language models are unsupervised multitask learners.
\newblock \emph{OpenAI blog}, 1\penalty0 (8):\penalty0 9, 2019.

\bibitem[Raviv et~al.(2022)Raviv, Lupyan, and Green]{RAVIV2022VariabilityCogSci}
Limor Raviv, Gary Lupyan, and Shawn~C. Green.
\newblock How variability shapes learning and generalization.
\newblock \emph{Trends in Cognitive Sciences}, 26\penalty0 (6):\penalty0 462--483, 2022.
\newblock ISSN 1364-6613.
\newblock URL \url{https://www.sciencedirect.com/science/article/pii/S1364661322000651}.

\bibitem[Regev(2005)]{Reg05}
Oded Regev.
\newblock {On Lattices, Learning with Errors, Random Linear Codes, and Cryptography}.
\newblock In \emph{Proc. of the ACM Symposium on Theory of Computing}, 2005.

\bibitem[Touvron \& et~al.(2023)Touvron and et~al.]{llama2touvron2023}
Hugo Touvron and et~al.
\newblock Llama 2: Open foundation and fine-tuned chat models.
\newblock \emph{arXiv preprint arXiv:2307.09288}, 2023.

\bibitem[Touvron et~al.(2023)Touvron, Lavril, Izacard, Martinet, Lachaux, Lacroix, Rozière, Goyal, Hambro, Azhar, Rodriguez, Joulin, Grave, and Lample]{llamatouvron2023}
Hugo Touvron, Thibaut Lavril, Gautier Izacard, Xavier Martinet, Marie-Anne Lachaux, Timothée Lacroix, Baptiste Rozière, Naman Goyal, Eric Hambro, Faisal Azhar, Aurelien Rodriguez, Armand Joulin, Edouard Grave, and Guillaume Lample.
\newblock Llama: Open and efficient foundation language models.
\newblock \emph{arXiv preprint arXiv:2302.13971}, 2023.

\bibitem[Vapnik \& Kotz(2006)Vapnik and Kotz]{vapnik82}
Vladimir Vapnik and S.~Kotz.
\newblock \emph{Estimation of Dependences Based on Empirical Data: Empirical Inference Science (Information Science and Statistics)}.
\newblock Springer-Verlag, Berlin, Heidelberg, 2006.
\newblock ISBN 0387308652.

\bibitem[Vaswani et~al.(2017)Vaswani, Shazeer, Parmar, Uszkoreit, Jones, Gomez, Kaiser, and Polosukhin]{vaswani2017attention}
Ashish Vaswani, Noam Shazeer, Niki Parmar, Jakob Uszkoreit, Llion Jones, Aidan~N Gomez, {\L}ukasz Kaiser, and Illia Polosukhin.
\newblock Attention is all you need.
\newblock In \emph{Advances in neural information processing systems}, pp.\  5998--6008, 2017.

\bibitem[Wang et~al.(2022)Wang, Chen, and Zhu]{WangCurriculumSurvey2021}
Xin Wang, Yudong Chen, and Wenwu Zhu.
\newblock A survey on curriculum learning.
\newblock \emph{IEEE Transactions on Pattern Analysis and Machine Intelligence}, 44\penalty0 (9):\penalty0 4555--4576, 2022.

\bibitem[Zhang et~al.(2017)Zhang, Bengio, Hardt, Recht, and Vinyals]{zhang2017understanding}
Chiyuan Zhang, Samy Bengio, Moritz Hardt, Benjamin Recht, and Oriol Vinyals.
\newblock Understanding deep learning requires rethinking generalization.
\newblock In \emph{International Conference on Learning Representations}, 2017.
\newblock URL \url{https://openreview.net/forum?id=Sy8gdB9xx}.

\bibitem[Zhong et~al.(2023)Zhong, Liu, Tegmark, and Andreas]{zhong2023pizza}
Ziqian Zhong, Ziming Liu, Max Tegmark, and Jacob Andreas.
\newblock The clock and the pizza: Two stories in mechanistic explanation of neural networks.
\newblock In \emph{Thirty-seventh Conference on Neural Information Processing Systems}, 2023.
\newblock URL \url{https://openreview.net/forum?id=S5wmbQc1We}.

\bibitem[Zipf(1935)]{zipf1935psycho}
GK~Zipf.
\newblock The psycho-biology of language: an introduction to dynamic philology.
\newblock 1935.

\end{thebibliography}
\bibliographystyle{ICLR2025Template/iclr2025_conference}

\newpage

\appendix
\section*{Appendix}

\section{Learning dynamics and overfitting in math transformers}\label{app:overfit}

\begin{figure} [h]
\center\small
\includegraphics[width=\textwidth]{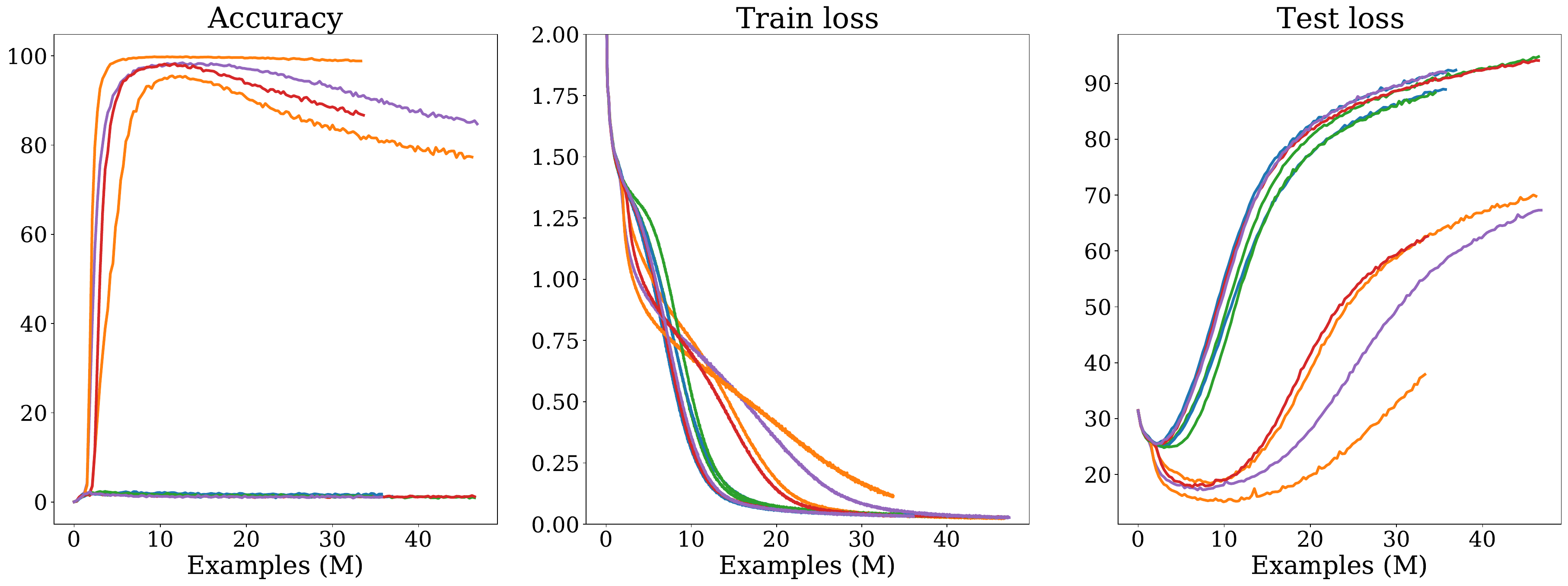}
  \caption{\small {{\bf Learning curves for eigenvalue computation of 5x5 matrices:} Accuracy, train and test loss, for $10$ models trained on a data budget of $200,000$, as a function of training budget (TB). The curves represent different seeds. Note the {\em initial phase}, common to all curves, up to a sharp transition of test loss at $\sim 2$M TB. At this point the dark curves begin to overfit (test loss increases) while the light curves undergo another drop in test loss that initiates the {\em learning phase}. }}
  \label{fig:overfit}
\end{figure}

To gain some understanding on the relation between repetition and overfitting, we delve deeper into the typical training dynamics in our mathematics problems with transformers. We study learning curves to shed light on the interplay between overfitting and relative size of data versus training budget. We focus on learning to compute the eigenvalues of $5\times 5$ symmetric 
matrices~\citep{charton2022linear} for illustrative purposes, but the observed dynamics are common to all our problems (e.g.~see Figure~\ref{fig:baseline} (Right)). Figure \ref{fig:overfit} illustrates training of $10$ models on a data budget of $200,000$ samples, with increasing training budget (up to $30$ million) resulting in increased repetition.
Learning curves exhibit a {\em step shape}, which gives rise to three phases:
\begin{itemize}[noitemsep,nolistsep]
\item {\em Initial phase:} training and test loss decrease (up to TB of about $2$M), accuracy remains low.
\item {\em Learning phase:} training and test loss drop suddenly, accuracy increases steeply from a few percents to $90\%$ (for the next $1$-$3$M of TB). This phase is absent for those models that overfit too early (dark curves in Figure \ref{fig:overfit}). 
\item {\em Saturation phase:} the model learns the remaining accuracy. 
\end{itemize}

Recall that we say that {\em overfitting} occurs when the test loss starts increasing while training loss continues decreasing. Here we see that for {\em all} models there is an initial flattening of test loss after $\sim2$M training examples (about $10$ repetitions of the data budget\footnote{Our runs on a range of small data budgets (up to $250$ thousand) show similar initial step shape of test loss at $10$-$12$ repetitions.}). Then, some models start overfitting already during the initial phase (the $6$ dark colored curves in Figure~\ref{fig:overfit}), and for those the learning phase never happens and accuracy plateaus at about $2\%$. On the other hand, for the other $4$ models the learning phase begins before overfitting sets in (the pale colored curves in Figure~\ref{fig:overfit}), the task is learned in full (to over $95\%$ accuracy), and overfitting is delayed until after that point. 
Eventually, these four models start to overfit at training budgets of about $10$ million examples, and a slight drop in accuracy is observed in some models (but not all), after $15$ million examples ($75$ epochs on the training set). We observe similar effects for different data budgets.

These experiments illustrate the relation between overfitting and learning. Once a model overfits, it stops learning, accuracy saturates, and eventually sometimes decreases.  On the other hand, once a model trained on limited data starts learning, overfitting is delayed by many more epochs.

\section{Additional figures and experiments for modular multiplication}\label{app:add_exp}

\begin{figure}[h]
\center \small
\includegraphics[width=0.5\textwidth]{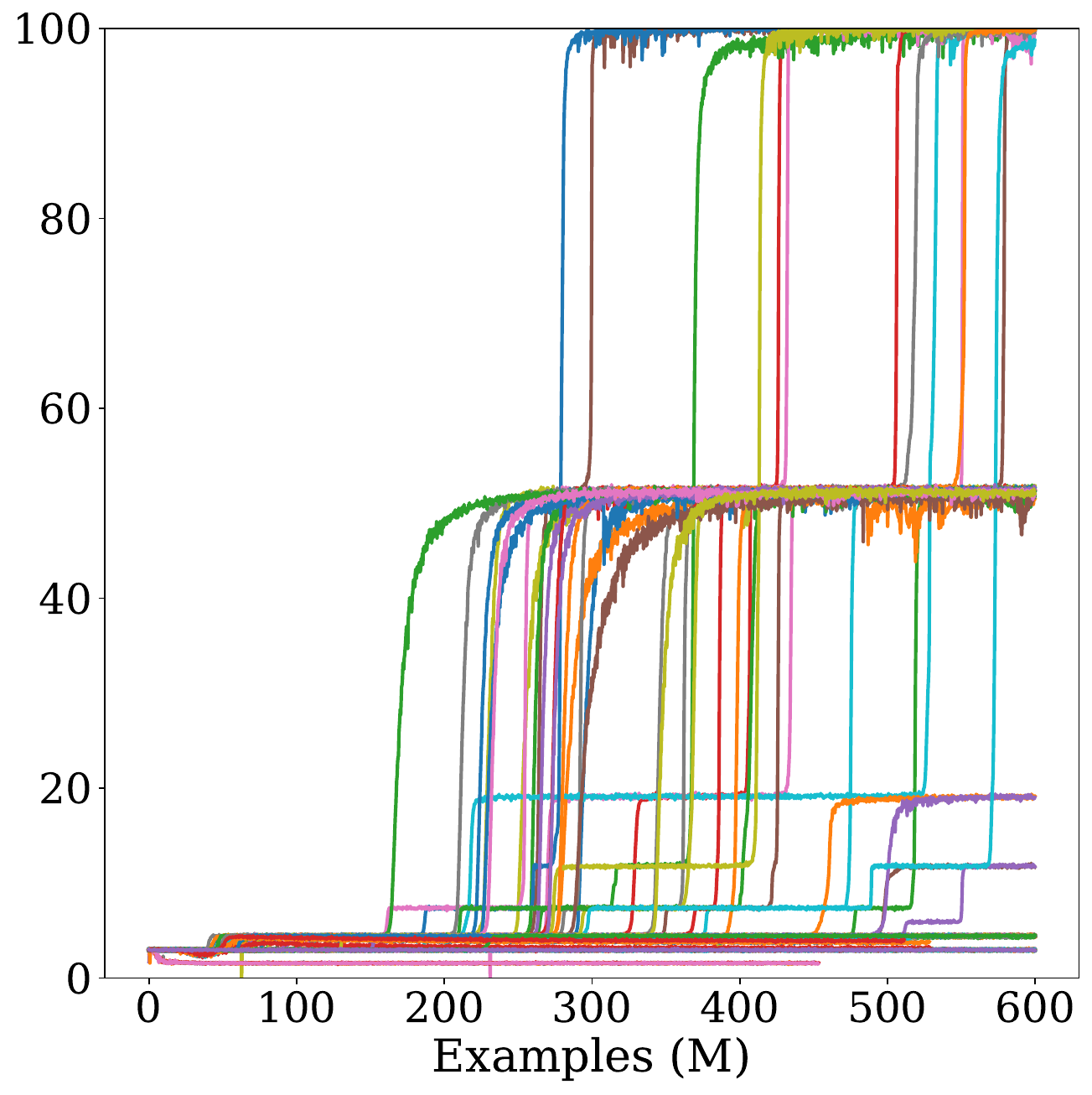}
  \caption{\small {\textbf{Learning curves for modular multiplication:} Test accuracy for different model initializations. We see a clear step-like learning curve with a plateau just above $50\%$ accuracy before jumping to near perfect accuracy.}}
  \label{fig:stepmodmul}
\end{figure}

Figure \ref{fig:stepmodmul} provides learning curves (test error) for modular multiplication, illustrating step-like learning, which motivates us to use the number of models achieving $50+\%$ resp. $99\%$ accuracy as our performance metric.

\begin{figure}[h]
\center \small
\includegraphics[width=0.8\textwidth]{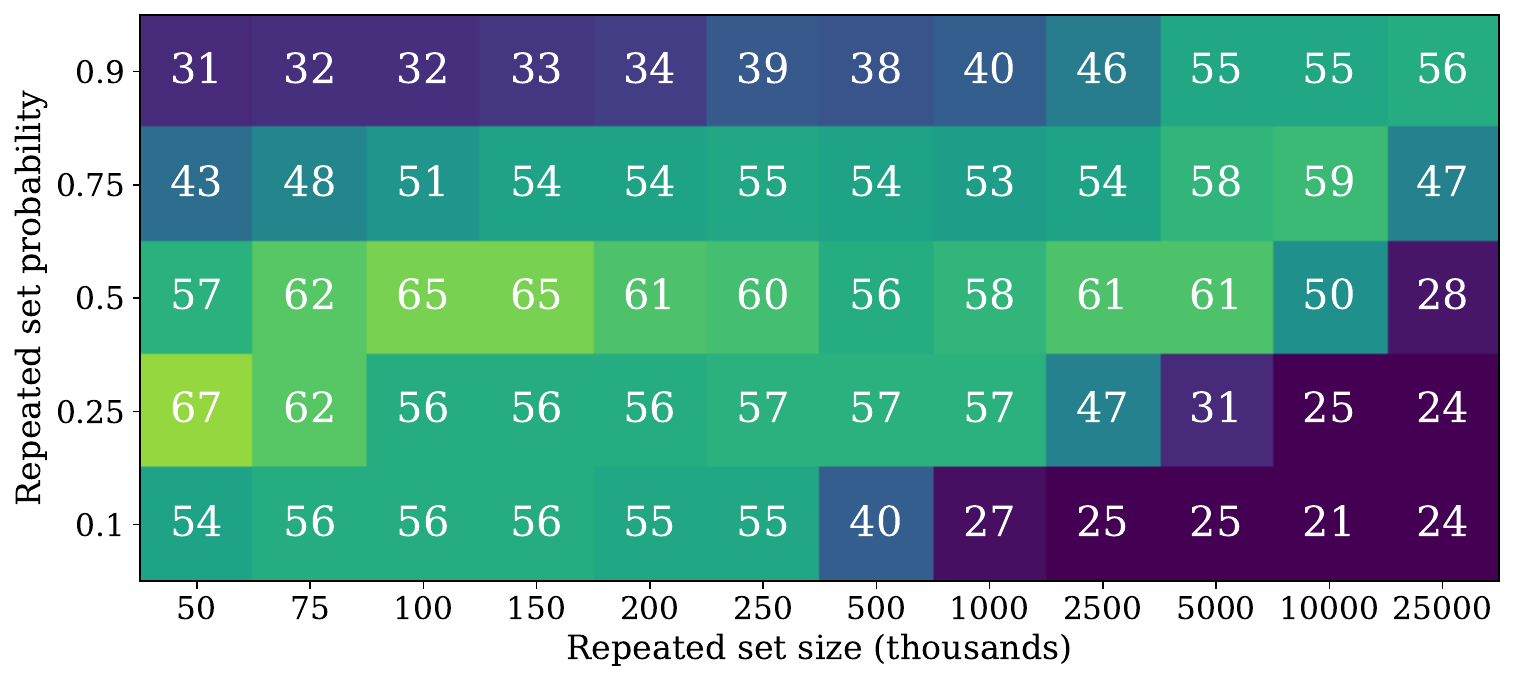}
  \caption{\small {\textbf{Two-set training for the GCD problem for $\infty$-models:} Number of correctly predicted GCD as a function of small set size $S$ and $p$, each averaged over $6$ models. Data budget {\em and} training budget equal $600$M ($\infty$-models). Note the high performance for very small sets $S$ of sizes between $50$ and $200$ thousand, with $p=0.25$ and $p=0.5$ compared to ``standard" training with the same data budget, predicting $25$ GCD correctly (see Section \ref{sec:repetition} }).}
  \label{fig:twosamples_inf}
\end{figure}

\begin{figure}[t]
\center \small
\includegraphics[width=0.45\textwidth]{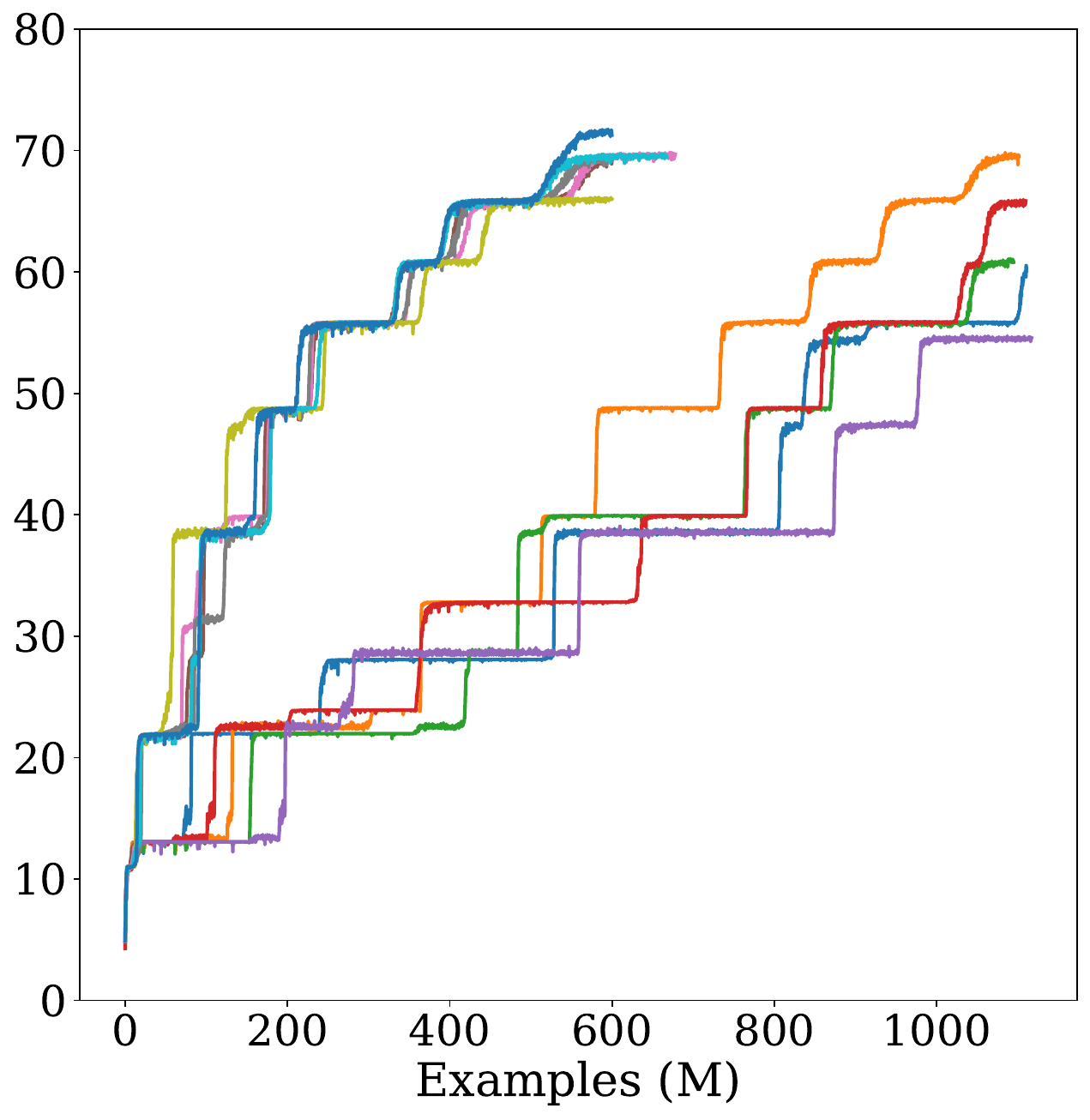}
  \caption{\small {\textbf{Two-set versus single-set training for the GCD problem:} Number of correctly predicted (test) GCD as a function of training budget (up to $1$B) and data budget of  $50$M  Two-set training with $p=0.25$ and $|S|=50,000$ (top 6 curves)  versus single-set training (lower $6$ curves). With enough TB, single-set training achieves comparable performance with two-set training.}}
  \label{fig:twosamples_25M2}
%  \vspace{-0.5cm}
\end{figure}

Figures \ref{fig:twosamples_inf} and \ref{fig:twosamples_25M2} as well as Tables \ref{tab:modmul_2samples} and \ref{tab:modmul_2samples_app} provide additional results for modular multiplication in the two-set setting. 

\begin{table}[h]
\small
\centering
\begin{tabular}{c|cc|cc|cc|cc}
\toprule
\textbf{$(p,S)$/ Data budget} & \multicolumn{2}{c|} {\textbf{$25$M}} &   \multicolumn{2}{c|}{\textbf{$50$M}} & \multicolumn{2}{c|}{\textbf{$100$M}} & \multicolumn{2}{c} {\textbf{$\infty$}} \\ 
& $>50\%$ & $99\%$ & $>50\%$ & $99\%$& $>50\%$ & $99\%$& $>50\%$ & $99\%$\\
\midrule
($0.1,500$K) & 2/10 &1/10 & 6/10 & 3/10 & 20/26 & 10/26 & {\bf 25/26} & 8/26\\
($0.1,1$M) & {\bf 5/10} & {\bf 5/10} & 8/10 & 4/10 & 22/26 & 6/26 & 0/26 & 0/26\\
($0.25,2.5$M) & 2/10 &1/10 & {\bf 9/10} & {\bf 5/10} & 20/26 & 9/26 &  24/26 & {\bf 15/26}\\
($0.25,5$M) & 3/10 &1/10 & 9/10 & 4/10 & {\bf 24/26} & 10/26 & 5/26 & 0/26\\
($0.5,10$M) & 3/10 &3/10 & 8/10 & 5/10 & 23/26 & {\bf 14/26} & 23/26 & 12/26\\
($0.75,25$M) & - & - & - & - & 23/26 & 10/26 & 20/26 & 14/26\\
\midrule
Single set & 13/25 & 6/25 & 22/25 & 7/25 & 0/30 & 0/30 & 0/30 & 0/30\\
\bottomrule
\end{tabular}
\caption{\small {\bf Two-set training on modular multiplication.} For a training budget of $600$M we show the number of models (random initializations) that achieve $50+\%$ and $90\%$ accuracy for several data budgets and sizes of the more frequent sets $S$, and probabilities $p$. The baseline of single-set traning from Section \ref{sec:repetition} is given in the last line. Similar results for training budgets of $300$M and $450$M are given in Table \ref{tab:modmul_2samples_app}. }
\label{tab:modmul_2samples}
%\vspace{-0.75cm}
\end{table}

\begin{table}[h!]
\centering
\resizebox{\columnwidth}{!}{%
\begin{tabular}{c|ccc|ccc|ccc|ccc}
\toprule
 & \multicolumn{6}{c|} {\textbf{Data budget $25$M}} &   \multicolumn{6}{c}{\textbf{ Data budget $50$M}} \\ 
 &  \multicolumn{3}{c}{$>50\%$} &  \multicolumn{3}{c|}{$99\%$} &  \multicolumn{3}{c}{$>50\%$} &  \multicolumn{3}{c}{$99\%$}\\
& $300$M & $450$M & $600$M & $300$M & $450$M & $600$M & $300$M & $450$M & $600$M & $300$M & $450$M & $600$M \\
\midrule
($0.1,500$K) & 1 & 2 & 2 & 0 & 1 & 1 & 4 & 5 & 6 & 0 & 1 & 3\\
($0.1,1$M) & 1 & {\bf 5} & {\bf 5} & 0 & {\bf 3} & {\bf 5} & 3 & 6 & 8 & 0 & 1 & 4 \\
($0.25,2.5$M) & 2 & 2 & 2 & 0 & 1 & 1 & 5 & {\bf 9} & {\bf 9} & 0 & 1 & {\bf 5}\\
($0.25,5$M) & {\bf 3} & 3 & 3 & 0 & 0 & 1 & 4 & {\bf 9} & {\bf 9} & 0 & 1 & 4\\
($0.5,10$M) & 2 & 3 & 3 & 0 & 2 & 3 & {\bf 7} & 7 & 8 & 0 & {\bf 2} & {\bf 5}\\
\midrule
Single set (/10) & 3.6 & 4.8 & 5.2 & 0.4 & 1.2 & 2.4  &  2.4 & 7.6 & 8.8 & 0 & 0.8 & 2.8\\
Single set (/25) & 9/25 & 12/25 & 13/25 & 1/25 & 3/25 & 6/25  &  6/25 & 19/25 & 22/25 & 0/25 & 2/25 & 7/25\\
\bottomrule
\end{tabular}%
}
\caption{\small {\bf Two-set training on modular multiplication.} For training budgets of $300$M, $450$M and $600$M we show the number of models out of 10 (random initializations) that achieve $50+\%$ and $90\%$ accuracy for data budgets $25$M and $50$M, and sizes of the more frequent sets $S$, and probabilities $p$. The baseline of single-set training is given in the last line, out of 25 models. The next to last line renormalizes this to out of 10. }
\label{tab:modmul_2samples_app}
\end{table}

\section{Ablation results}\label{app:ablation}

\subsection{Curating the small sample}

In two-set training, the examples in the small set are chosen at random from the overall training set. In this section, we experiment with curating the small set, by {\em selecting} the examples that will be repeated during training. As in curriculum learning, selecting easier or more informative examples may help improve performance. Perhaps when increasing the frequency of our small random set, what really matters is the repetition of some particular examples, rather than all?
The GCD problem is particularly well suited for this type of investigation, due to the inverse polynomial distribution of outcomes ($\text{Prob}(\text{GCD}=k)\sim \frac{1}{k^2}$).
On this problem, we leverage the findings of \citet{charton2023gcd}, who observes that $\infty$-models trained from log-uniform distributions of inputs and/or outcomes ($\text{Prob}(\text{GCD}=k)\sim \frac{1}{k}$) learn better. 

We experiment with four settings of $|S|$ and $p$, which correspond to the best results in our previous experiments (Section \ref{sec:2-set}): $50,000$ and $150,000$ with $p=0.25$ and $150,000$ and $500,000$ with $p=0.5$, for a data budget of $100$ million and training budget of $600$M. For every setting, we train $5$ models with the following three choices for $S$: log-uniform inputs, uniform GCD or both log-uniform inputs and GCD. We use two-set training with a random small set $S$ as our baseline. Table~\ref{tab:gcd_xp3} shows that the performance of models using log-uniform inputs, or uniform GCD, is slightly lower than the baseline. Models trained on log-uniform inputs and GCD achieve slightly better performance, but we note that  models trained on the small set distribution only ($p=1$) would predict $91$ GCD. On these three distributions, curating the small set proves disappointing.

In curriculum learning fashion, we also experiment with small sets $S$ of a few  ``easier cases'': small inputs (from $1$ to $1000$), GCD that are products of $2$ and $5$, the easiest to learn in base $1000$ \citep{charton2023gcd}, and GCD between $1$ and $10$ (the most common outcomes). We observe that while models trained with  small inputs in $S$ perform on par with the baseline, models trained on ``easy GCD'' perform slightly worse.

Finally, inspired by arguments that rare tail outcomes might require particular attention for learning 
\citep{dohmatob2024aTaleTails}, we experiment with small sets composed of examples from the tail of the training distribution, namely, large GCD. \citet{charton2023gcd} observes that these are both harder to learn, and less common in the training set. Specifically, we create $S$ with examples with GCD larger than $k$ (for $k$ ranging from $1$ to $5$). While experiments achieve the best accuracies compared to the other curation schemes we proposed, and values of $k$ equal to $2$ and $3$ train slightly faster, they remain a little below the baseline both in accuracy and learning speed.

\begin{table} [h]
\centering
\small
\begin{tabular}{lccccc}
\toprule
& & & & & Training budget \\
& 50k / 0.25 & 150k / 0.25 & 150k / 0.5 & 500K / 0.5 & for 60 GCD (M)\\
\midrule
Log-uniform inputs & 55.9 & 59.4 & 57.9 & 62.0 & 332\\
Uniform GCD & 55.9 & 54.5 & 41.9 & 54.9 & - \\
Log-uniform inputs and GCD & 62.2 & {\bf71.7} & {\bf 66.5} & {\bf 72.6} & 88\\
\midrule
Small inputs (1-1000) & 61.2 & {\bf 67.5} & 62.6 & 62.9 & 247\\
GCD 1- 10 & 59.9 & 63.8 & 55.8 & 62.3 & 401\\
GCD products of 2 and 5 & 54.2 & 39.8 & 40.7 & 30.1 & 548\\
\midrule
All GCD but 1 & {\bf 65.4} & 63.7 & 56.7 & 58.1 & 405\\
All GCD but 1,2 & {\bf 66.8} & 60.0 & 62.8 & 56.9 &326\\
All GCD but 1,2,3 & {\bf 66.7} & 58.4 & 62.8 & 58.2  & 327\\
All GCD but 1,2,3,4 & {\bf 65.5} & 60.3 & 62.8 & 56.9 & 379\\
All GCD but 1,2,3,4,5 & {\bf 66.5} & 60.6 & 64.9 & 56.3 & 376\\
\midrule
GCD product of 2, 3, and 5 & {\bf 66.1} & 59.4 & 59.8 & 47.3 & 359 \\
Prime GCD & 64.9 & 62.5 & 58.8 & 64.7 & 422\\
GCD divisible by primes $\geq$ 11& 60.1 & 54.4 & 35.7 & 42.7 & 569 \\
\midrule
Baseline (two-set training) & {\bf 69.4} & 61.9 & {\bf 65.9} & 59.4 & 373\\
\bottomrule
\end{tabular}
\caption{\small \textbf{GCD problem: cherry-picking the small set}. (Left) Number of (test) GCD predicted for training budget of 600 million examples, average of 5 models (3 models for baseline). \textbf{bold}: more than 65 GCD predicted. (Right) Training budget needed to predict 60 GCD, fastest of 20 models (of 12 models for baseline).}
\label{tab:gcd_xp3}
\end{table}

Overall, these experiments suggest that in two-set training, random selection of the small set may be optimal. Selecting a small set of easy cases (GCD multiple of $2$ and $5$), and examples that are known to help training (log-uniform inputs) does not help, and limiting the small set to edge cases from the tail of the outcome distribution brings no improvement to performance. This is a counter-intuitive, but significant result.

\subsection{Batching in two-set training: mixed batches are needed}\label{app:monobatch}

In all experiments, during training, the model computes gradients over minibatches of $64$ examples. In two-set training, minibatches mix examples from the small and large set. We experimented with using ``mono-batches" that use samples from one set at a time. For instance, when training with $p=0.25$, $25\%$ of minibatches would use examples from the small set (of size $S$) only, and $75\%$ would only use those from its complement. 

On the {\bf GCD problem}, we rerun the most successful two-set experiments (Section \ref{sec:2-set}) with ``mono-batches" for 
$S=50$K, $100$K and $250$K, and $p=0.25$ and $0.5$. For training budgets of $600$M and data budget of $100$M examples, the models trained on mixed batches predicted $62$ to $69$ GCD (Section \ref{sec:2-set}). With ``mono-batches", the number of correctly predicted GCD never rises above $15$. For {\bf modular multiplication}, we experimented with the following $(S,p)$ pairs ($S$ in millions): $(0.5,0.1), (2.5,0.25)$ and $(10, 0.5)$ with data budget $100$M and training budget $600$M. With these settings, mixed-batch models achieve an average accuracy of $67\%$ or more (Section \ref{sec:2-set}).  With ``mono-batches", none of the models manages to learn (accuracy around $4\%$). This indicates that {\bf mixed batching of samples from each of the two sets plays a central role for the two-set effect}.% and preventing early overfitting. 

\subsection{Shifting the small set}\label{app:shift}

In these experiments, we study, in two-set training, the possible impact of overfitting on the small set, by refreshing the small set with fresh examples periodically. This mimics certain aspects of curriculum learning, where the training set is changed over time. On the GCD experiments, with a data budget of $100$ million, a training budget of $600$ million, we shift the small set as training proceeds, so that examples in the small set are seen $k$ times on average. At the beginning of training, the small set is the $S$ first elements in the train set. After training on $kS/p$ examples, examples in the small set have been seen $k$ times, and the small set is shifted to elements $S+1$ to $2S$ of the training set. 

Table~\ref{tab:shift} provides performances for two-set training with shift, for different values of $p$, $S$ and $k$, for a data budget of $100$ million, and a training budget of $600$ million. It is interesting to note that shifting brings no improvement to 2-set training.

\begin{table} [h]
\centering
\small
\begin{tabular}{l|cccc|cccc|cccc}
\toprule
$S$ & \multicolumn{4}{c|}{250,000} & \multicolumn{4}{c|}{500,000} & \multicolumn{4}{c}{1,000,000} \\
k & 10 & 25 & 50 & 100 & 10 & 25 & 50 & 100 & 10 & 25 & 50 & 100 \\
\midrule
$p=1.0$ & 37 & 22 & 21 & 22 & 37 & 38 & 30 & 31 & 55 & 45 & 37 & 30\\
$p=0.9$ & 47 & 38 & 38 & 38  & 55 & 47 & 43 & 39 & 55 & 48 & 47 & 47 \\
$p=0.75$ & 56 & 38 & 54 & 48 & 56 & 55 & 49 & 55 & 60 & 56 & 55 & 56 \\
$p=0.5$ & 61 & 56 & 56 & 58 & 61 & 60 & 56 & 58 & 64 & 63 & 63 & 61\\
$p=0.25$ & 56 & 62 & 61  & 63 & 49 & 63 & 63 & 61 & 49 & 63 & 62 & 63\\
\bottomrule
\end{tabular}
\caption{\small \textbf{Shifted two-set training.} GCD predicted, average of 3 models, trained on a budget of 600 millions, and a data budget of 100 million, for different values of S, p and k.}
\label{tab:shift}
\end{table}

\subsection{From two-set to many-set training}\label{app:manyset}

Two-set training with a small randomly selected subset $S$ amounts to assigning different probabilities to elements in the training set. For a randomly shuffled training set of size $N$, two-set training amounts to selecting the first $S$ elements with probability $p/S$ (with replacement) and the $N-S$ last with probability $(1-p)/(N-S)$, a step-function distribution over $\{1,\dots,N\}$.
We now generalize this approach by introducing a probability law $P$ such that $P(i)$ is the probability of selecting the $i$-th example in the training set. Our motivation is to obtain a smooth, possibly more principled, distribution than the step-function induced by the two-set approach. Pragmatically, a one-parameter family of smooth distributions eliminates the need to tune both $S$ and $p$. Lastly, we can study whether a smooth decay in frequency might be even more beneficial than a non-continuous two-set partition.

In this section, we consider a discrete exponential distribution:  $$P(i) \sim \beta e^{-\beta i / N},$$ with $\beta > 0$, suitably normalized\footnote{The normalization factor is $(1-e^{-\beta})^{-1}$. In our calculations we will approximate it by $1$ to simplify computing  $S_{\text{eff}}$. For the range of $\beta$ we consider, the resulting approximation error is negligible. In general, for fixed $p$, to compute the size of the set $S(p)$ of first elements that carry probability mass $p$, we can use 
 $\beta \approx - \ln{(1-p)}N/|S(p)|$. }. If $\beta$ tends to $0$, $P$ tends to the uniform distribution, and implements the single-set strategy of Section~\ref{sec:repetition}. As $\beta$ becomes large, a small fraction of the full training set is sampled ($99\%$ of the probability mass lies on the $4.6N/\beta$ first elements, $99.99\%$ on the first $9.2N/\beta$). For intermediate values of $\beta$, the model oversamples the first elements in the training set, and undersamples the last: we have a continuous version of two-set training. 
To allow for comparison with two-set training, we define $S_{\text{eff}}$ such that the first $S_{\text{eff}}$ examples in the training set jointly are sampled with
 probability $25\%$. In this setting, $10\%$ of the probability mass is on the $0.37S_{\text{eff}}$ first training examples, and $99\%$ on the first $16S_{\text{eff}}$. 

For GCD, we experiment with values of $\beta$ ranging from $5.8$ to $1152$ ($S_{\text{eff}}$ from 25,000 to $5$ million)\footnote{Note that for these values of $\beta$ the distinction between DB $100$M and unlimited DB becomes essentially meaningless, as the tails of the training set are sampled exceedingly rarely.}. Table~\ref{tab:gcd_xp4_main} shows that for our training budget of $600$ million examples, the best model ($S_{\text{eff}}=3$M) predicts $65$ correct GCD, slightly less than what was achieved with two-set training (Section \ref{sec:2-set}). 

For modular multiplication, we need lower $\beta$ (i.e larger $S_{\text{eff}}$) for our training budget of $600$M. We report the number of models (out of 25 for each setting) that learn to accuracy above $50\%$ and $95\%$ respectively (Table~\ref{tab:modmul_xp4}). Again we see that these results are comparable to two-set training (Section \ref{sec:2-set}). 

\begin{table} [h]
\centering
\small
\begin{tabular}{lcccccccc}
\toprule
$S_{\text{eff}}$ & 2.5M &  5M & 6M & 8M & 10M & 12M & 14M \\
$\beta$ & 11.5 & 5.8 & 4.8 & 3.6 & 2.9 & 2.4 & 2.1 \\
\midrule
\# Models with $95\%$ accuracy & 2 & 9 & 11 & 13 & 7 & 4 & 3\\
\# Models with $50\%$ accuracy & 4 & 16 & 25 & 22 & 17 & 13 & 6\\
\bottomrule
\end{tabular}
\caption{\small \textbf{Modular multiplication with different exponential distributions.} 25 models trained on 600 million examples.}
\label{tab:modmul_xp4}
\end{table}

We conclude that the benefits observed in two-set training do not pertain to the specific two-set partition of the training set; rather, it seems that the core of the effect lies in the non-uniform sampling frequency distribution over the (randomly ordered) training set, with a range of frequencies.

\subsection{Varying the optimizer}\label{app:optimizers}

Some effects observed in deep learning depend on the optimizer, with grokking being a prominent example  \citep{Power2022Grokking}. Here we provide experimental evidence to show that our findings hold for a variety of optimizers and are thus {\em robust} and {\em universal}. We rerun models used for the GCD problem with different optimizers. Specifically, we trained models to predict GCD, with a training budget of $600$ million examples, single and two-set training (with $|S|=50,000$ and $p=0.25$), and data budgets of $25$ million, $50$ million and unlimited. We considered four optimizer settings: 
\begin{itemize}[nosep]
\item Adam without dropout or weight decay,
\item Adam with weight decay $0.01$,
\item Adam with dropout ($0.1$) in the feed-forward networks of the transformer,
\item AdamW with weight decay $0.01$.
\end{itemize}

Table~\ref{tab:optim} presents the best performance of $5$ models for each configuration. On average, dropout has an adverse effect on learning, but there is no clear benefit of using weight decay, or AdamW over Adam. Importantly, the separation in performance between single-epoch unlimited training, training on smaller data budgets with more repetitions and two-set training persists across optimizers: the effects we present are robust.

\begin{table} [h]
\centering
\small
\begin{tabular}{l|ccc|ccc}
\toprule
& \multicolumn{3}{c|}{One-set} & \multicolumn{3}{c}{Two-set}\\
& Unlimited & 50M & 25M & Unlimited & 50M & 25M \\
\midrule
Adam & 28 & 49 & 61 & 70 & 72 & 63 \\
Adam wd=0.01 & 30 & 56 & 61 & 70 & 70 & 66 \\
AdamW wd=0.01 & 29 & 50 & 58 & 69 & 72 & 67 \\
Adam dropout=0.1 & 24 & 40 &49 & 66 & 66 & 66\\
\bottomrule
\end{tabular}
\caption{\small \textbf{Modular multiplication with different optimizers.} Correctly predicted GCD of the best (of 5) models for various optimizers. The effects we observe are robust under change of optimizer, with a very small degradation for dropout for both the unlimited (single-epoch) and limited DB. }
\label{tab:optim}
\end{table}

\end{document}